\documentclass[journal]{IEEEtran}

\usepackage[utf8]{inputenc}
\usepackage[T1]{fontenc}
\usepackage{lmodern}
\usepackage{graphicx}
\usepackage{cite}
\usepackage{amsmath,amssymb}
\usepackage{booktabs}
\usepackage{xcolor}
\usepackage{hyperref}
\usepackage{algorithm}
\usepackage{algpseudocode}
\usepackage{tabularx}
\usepackage{array}
\usepackage{placeins}              

\title{Leak-Free Cross-Validated Stacking with Per-Architecture
Calibration for Sand-Boil Segmentation in Earthen Levees}

\author{%
Padam Jung Thapa\textsuperscript{1},
Anav Katwal\textsuperscript{2,\$},
Ayon Dey\textsuperscript{2,\$},
Abdullah Bin Naeem\textsuperscript{2,\$},
Steve Sloan\textsuperscript{3},
Kendall Niles\textsuperscript{3},
Md Tamjidul Hoque\textsuperscript{2,*}\\[4pt]
{\normalsize
\textsuperscript{1}The Center for Advanced Computer Studies (CACS), School of Computing and Informatics,\\
University of Louisiana at Lafayette, Lafayette, LA 70504, USA\\
\textsuperscript{2}Department of Computer Science, LSU New Orleans, New Orleans, LA 70148, USA\\
\textsuperscript{3}US Army Corps of Engineers, Engineer Research and Development Center, Vicksburg, MS 39180, USA\\[3pt]
Emails: padam-jung.thapa1@louisiana.edu;\quad
\{akatwal, adey, anaeem\}@lsuneworleans.edu;\quad
\{steven.d.sloan, kendall.n.niles\}@erdc.dren.mil;\quad
thoque@lsuneworleans.edu\\[1pt]
\textsuperscript{\$}Equal contribution.\quad
\textsuperscript{*}Corresponding author: thoque@lsuneworleans.edu}%
}

\begin{document}
\maketitle
\bstctlcite{BSTcontrol}

\begin{abstract}
Sand boils, points where water seeping beneath an earthen levee
re-emerges at the surface, are early warnings of internal erosion, and
deep segmentation networks are increasingly used to find them in
inspection photographs. Annotated examples are scarce, and two common
ways of working around that scarcity quietly inflate reported accuracy:
tuning ensemble weights on the same images later used to score them, and
training on synthetic images derived from the very photographs held out
for testing. We present a sand-boil segmentation framework that closes
both loopholes. Every synthetic image carries a pointer to its real
parent, and a per-fold filter excludes any image whose parent is held
out; five encoder--decoder backbones are trained under five-fold
cross-validation, calibrated by one temperature scalar each, and
combined by a per-pixel meta-learner fitted only on out-of-fold
predictions. On the held-out test set the proposed Updated SandBoilNet
reaches an intersection-over-union of $0.707$ over three seeds, against
$0.608$ for the published original re-evaluated on the same split. Under
the stacking protocol the calibrated stack reaches $0.681$ against
$0.694$ for the strongest fold-averaged member, so it does not improve on
the best single model; eight meta-learner families reproduce that
outcome, which we trace to a mean pairwise error correlation of $0.894$
among members. A synthetic pool filtered for label fidelity lifts the
champion to $0.718$ over three seeds against a $0.707$ control. We also introduce a mask-conditioned synthesis route
that makes the conditioning mask the label by construction, giving
labelled training images at zero annotation cost.
\end{abstract}


\section{Introduction}
\label{sec:intro}

Pixel-level segmentation of sand boils from levee-inspection imagery
is the most direct lever for proactive flood-risk management on
earthen-levee networks. Sand boils form when water under hydraulic
pressure transports fine sand upward through a permeable foundation
layer, producing a small saturated dome at the surface; the
mechanism is the early visible stage of internal erosion, which, if
left untreated, can progress to a sudden breach during a high-water
event \cite{ilit2006katrina}. Modern automated work on this class is
built on convolutional encoder--decoders, of which Panta
\textit{et al.}'s SandBoilNet \cite{panta2023sandboilnet} is the
domain-specific state of the art, reporting an
intersection-over-union of $0.5743$ and a balanced accuracy of
$0.8552$ on a held-out real test set. Re-evaluated in our own
pipeline on this paper's held-out test split, the published
checkpoint reaches $0.608$ IoU, which we use as the like-for-like
reference throughout. The curated dataset used in
the present study (whose size is specified in
Section~\ref{subsec:dataset}) is, to our knowledge, one of the
larger sets of its kind, and is the same dataset SandBoilNet was
originally trained on.

A natural way to extract residual accuracy beyond a single strong
backbone is to combine several backbones into an ensemble. Two
related ensembling strategies appear in the levee-defect literature:
pixel-level majority voting and weighted probability averaging
\cite{thapa2025thesis}. Both are easy to implement but share two
structural defects that this paper addresses directly; a related third
problem, false positives on visually confusable non-defects, is a property
of the segmenter rather than the ensemble and motivates our hard-negative
phase.

\paragraph{Problem 1: Subtle leakage in weighted-average ensembles.}
A weighted-average ensemble must choose a weight per base learner.
The natural way to choose those weights is to evaluate each base
learner on a held-out validation split, fit weights that minimise
the ensemble loss on that split, and report the ensemble's accuracy
on the same split. The procedure is convenient, but it tunes the
ensemble on the very split it then claims as evaluation. The
resulting numbers are inflated relative to a fair evaluation. The
cleanest fix is the cross-validated stacking framework of Wolpert
\cite{wolpert1992stacking}: train each base learner $K$ times under
a $K$-fold partition, collect each base learner's predictions on
the held-out fold only, and fit the meta-learner on the resulting
out-of-fold cube. Every prediction in that cube was produced by a
model that did not see the corresponding example during training,
so the meta-learner's weights themselves are unbiased.

\paragraph{Problem 2: A second leakage path through synthetic
descendants.} When synthetic data is mixed into the training set,
a new leakage path appears that has no equivalent in the
real-only regime. A synthetic image generated from a real parent
$x$ inherits some of $x$'s lighting, texture, and rim geometry,
even if the rendered scene is otherwise unrecognisable. If $x$
later lands in a held-out fold, training the ensemble on
synthetic descendants of $x$ leaks information about $x$ into the
training pool of the very fold that is supposed to be evaluating
on it. The fix is also structural: every synthetic record must
carry a pointer to its real parent, and the fold-construction
machinery must filter out synthetic descendants of every held-out
parent. We adopt this filter as a one-line set operation per fold
(Section~\ref{subsec:leak_free}) and treat it as an artefact-level
discipline that any pipeline mixing synthetic and real data should
adopt under cross-validation.

\paragraph{Problem 3: False positives on visually confusable
non-defects.} A segmenter trained on positive sand-boil imagery
alone tends to over-fire on visually similar non-defects (puddles
with sediment rings, hoof-print puddles, mole and gopher mounds,
dried-mud polygons, half-buried cobbles, algal mats, and so on)
that share dome-like or annular structure with a real sand boil.
Panta \textit{et al.}\ \cite{panta2023sandboilnet} address this
with a second training phase that exposes the segmenter to
defect-free levee imagery drawn from the same inspection archive.
We extend that protocol with two changes. First, on top of the
real defect-free pool we generate a synthetic hard-negative pool
spanning a curated catalogue of visually confusable categories,
produced by a defect-disabled production preset of the upstream
diffusion pipeline \cite{thapa2026diffusion}. Second, each synthetic negative is scored by
the public SandBoilNet baseline and weighted into the negative
phase in proportion to how strongly it fools the baseline. This
converts the negative phase from a static second pass into a
confusion-driven training loop in which the most informative
synthetic negatives receive the largest loss weight.

\paragraph{Contributions.} This paper makes two empirical
contributions and four methodological ones:
\begin{itemize}
\item An Updated SandBoilNet (ConvNeXt-S encoder, U-Net decoder,
spatial-and-channel squeeze-and-excitation (scSE) skip attention)
reaching $0.707$ IoU against $0.608$ for the published original
re-evaluated on the same split. Neither heavier attention variants nor a
four-fold resolution increase improve on it: the modernised encoder, not
added attention capacity, moves accuracy here.

\item A controlled study of ensembling and synthetic data in this
scarce-data regime. The calibrated stack reaches $0.681$ against $0.694$
for the best single model, an outcome eight meta-learner families
reproduce and which we trace to member correlation; synthetic
augmentation pays when the pool is filtered for label fidelity; and
neither a hard-negative phase nor higher input resolution improves on the
champion. Establishing which additions transfer to a scarce-data defect,
and which do not, is itself useful guidance.

\item A leak-free cross-validated stacking \emph{protocol} whose ensemble
accuracy is unbiased by construction. Its contribution is the
trustworthiness of the measurement rather than the size of a gain: here
that measurement returns no ensemble gain at all, a conclusion the leaky
protocols it replaces cannot reach, since their headline numbers conflate
an ensemble's benefit with the tuning that produced it.

\item A parent-image fold filter closing the silent leakage path through
synthetic descendants: each synthetic record carries a
\texttt{source\_image\_id} pointer to its real parent, and the
fold-construction driver excludes every descendant of a held-out parent
in $O(1)$.

\item A confusion-score-mined synthetic hard-negative phase, in which a
defect-disabled generator preset produces visually confusable negatives
and the sampler weights each in proportion to how strongly it fools the
public baseline.

\item A mask-conditioned synthesis method, \textsc{MaskCN}
(Section~\ref{subsec:maskcn}), that renders a fresh sand boil into a
\emph{target} mask, so the conditioning mask is the ground-truth label by
construction. It supplies fully labelled images at zero annotation cost,
has no real parent to leak, and adds procedural multi-boil fields the
real set cannot provide.
\end{itemize}

\paragraph{Scope.} The present paper is the second in a three-paper
series. The diffusion-based pipeline that produces both the
augmented training imagery and the synthetic hard-negative imagery
consumed here is the subject of the companion paper \cite{thapa2026diffusion}; it is treated
in Section~\ref{subsec:data_sources} only as an upstream module
that emits parent-tagged synthetic records. The deployment system,
human-in-the-loop correction interface, and curated dataset release
are the subject of a third paper in the series.

The remainder of the paper is organised as follows.
Section~\ref{sec:related_work} positions the contribution against
the literature on segmentation architectures for limited-data
regimes, stacked generalisation, calibration, and hard-negative
mining. Section~\ref{sec:methodology} develops the framework.
Section~\ref{sec:experiments} describes the experimental protocol.
Section~\ref{sec:results} reports quantitative and qualitative
results. Section~\ref{sec:discussion} examines the design choices
and remaining failure modes. Section~\ref{sec:conclusion}
concludes.

\section{Related Work}
\label{sec:related_work}

The relevant literature spans five threads: sand-boil and levee-defect
segmentation; encoder--decoder architectures for limited-data regimes;
synthetic-data augmentation; ensemble learning with stacked
generalisation and calibration; and hard-negative mining. The generative
machinery behind the synthetic data is covered only to the depth the
segmentation framework requires, a fuller account belonging in the
companion generation paper \cite{thapa2026diffusion}.

\subsection{Sand-boil and levee-defect segmentation}
\label{subsec:rw_levee}

Early automated levee-defect detection relied on edge filters,
thresholding, and handcrafted features fed into shallow classifiers
\cite{guan2023pavement}, which fared poorly under the lighting
variation, mud, vegetation, and reflective standing water typical of
in-field inspection photographs. Modern work is built on convolutional
encoder--decoders. Panta \textit{et al.}\ introduced IterLUNet
\cite{panta2023iterlunet} for levee crack segmentation and the more
directly relevant SandBoilNet \cite{panta2023sandboilnet}, an
encoder--decoder on a partially fine-tuned ResNet-50v2 backbone whose
skip connections carry PCA-based Channel and Spatial Attention with
Inception blocks. Its reported intersection-over-union of $0.5743$,
balanced accuracy of $0.8552$, and macro-averaged F1 of $0.7312$ provide
the principal point of comparison here. The same group later applied
related ideas to seepage detection
\cite{panta2024seepage,abdelguerfi2024erdc} and to imbalance-aware
culvert-and-sewer defect segmentation \cite{alshawi2024efpn}, and the
same family extends to marine oil-spill delineation
\cite{elsheref2026oilspillnet}. The field has since moved toward
UAV-borne inspection, with thermal-infrared and visible-light imagery
feeding detectors for seepage and embankment piping at field scale
\cite{hu2024manualtouav,duan2025uavpiping} and frequency-domain and
pruned-attention networks pushing toward real-time edge deployment
\cite{wang2025embfreqnet,wu2025prunedunet}. Those works target detection
or coarse localisation from aerial platforms, whereas we address
pixel-level segmentation of close-range photographs. Kuchi \textit{et
al.}\ \cite{kuchi2019sandboil} first paired a small synthetic sand-boil
set with a CNN classifier, but operated at image level. Neighbouring
work includes sinkhole detection with a depthwise separable U-Net
\cite{alshawi2024sinkhole}, while synthetic-aperture radar
\cite{james2011sar} offers insufficient resolution for features this
small. No prior sand-boil study combines its models through a leak-free
cross-validated stacking ensemble; our earlier work
\cite{thapa2025thesis} prototyped a weighted-average ensemble whose
weights were tuned on the split later used to report its accuracy, a
mild leakage we address in Section~\ref{subsec:disc_stacking}.

\subsection{Segmentation architectures for limited-data regimes}
\label{subsec:rw_segmentation}

Semantic segmentation assigns a class label to every pixel
\cite{garcia2018semseg}, and current architectures build on Fully
Convolutional Networks \cite{long2015fcn} by replacing dense classifier
heads with convolutional decoders. Common variants include PSPNet
\cite{zhao2017pspnet}, SegNet \cite{badrinarayanan2017segnet}, and
DeepLabv3 \cite{chen2017deeplabv3}. For the small-dataset regime
characteristic of infrastructure-defect imagery the U-Net family
\cite{ronneberger2015unet} remains dominant, since its skip connections
preserve the narrow boundary features that delimit a boil's rim. Notable
extensions include MultiResUNet \cite{ibtehaz2020multiresunet}, the
nested U-Net++ \cite{zhou2018unetpp}, and Attention U-Net
\cite{oktay2018attention}. The five backbones evaluated here are
deliberately not confined to that family: an architecture search retains
the best of five complementary families, comprising an Updated
SandBoilNet, a vision transformer, a nested convolutional U-Net, a
multi-scale pyramid network, and an atrous network
(Section~\ref{subsec:archs}), chosen because their differing inductive
biases make errors independent enough to stack while a shared
input--output geometry keeps their out-of-fold predictions combinable.
The wider defect-segmentation field is in parallel migrating from
convolutional encoder--decoders such as DeepCrack
\cite{liu2019deepcrack} toward transformer and hybrid designs
\cite{liu2023crackformer,cheng2022mask2former} and
toward prompting vision foundation models
\cite{ge2023cracksam,ahmadi2023sam}, a shift enabled by consolidated
benchmarks
\cite{kulkarni2022crackseg9k,benz2024omnicrack30k}.
Surveys of segmentation under scarce annotation
\cite{catalano2023fssreview,jiao2023limitedannotations} confirm that
strong augmentation and regularisation, rather than ever-larger
backbones, are the dominant levers in this regime.

\subsection{Generative synthetic data augmentation}
\label{subsec:rw_synth}

Expanding scarce defect datasets with generated imagery has progressed
from GAN-based synthesis \cite{zhang2021datasetgan,mao2023cgancrack} to
diffusion models that jointly emit images and pixel-aligned masks
\cite{wu2023diffumask,schnell2023scribblegen}. The
crucial design axis is where the label comes from. One family
\emph{infers} a mask for each generated image, by thresholding an
attention map \cite{wu2023diffumask} or a separately trained segmenter,
which reintroduces annotation noise and, when the generator is
conditioned on a real seed, a parent linkage. A second family
\emph{conditions} the generator on a target mask so that the mask is the
label by construction, made practical at high resolution by ControlNet
and related adapters \cite{zhang2023controlnet} and followed by
scribble-conditioned diffusion \cite{schnell2023scribblegen}. Our MaskCN
route (Section~\ref{subsec:maskcn}) sits in this second family: a
mask-ControlNet trained jointly with a sand-boil DreamBooth low-rank
adaptation \cite{ruiz2023dreambooth,hu2022lora} renders a fresh boil
into a prescribed mask, so the conditioning mask is the annotation at
zero labelling cost and, inheriting no real seed, has no parent linkage
to leak. In the defect domain, LoRA-fine-tuned latent diffusion lifts
segmentation mIoU on steel-surface inspection
\cite{leinena2024latentdiffusion}, classifier-side studies report that
diffusion-augmented training can approach or exceed real-only baselines
\cite{azizi2023synthetic,trabucco2024dafusion}, and
mask-preserving augmentations such as copy-paste remain strong, simple
baselines \cite{ghiasi2021copypaste,yun2019cutmix}. A recurring caution,
central to this paper, is that naive use of synthetic data inflates
reported accuracy through a subtle leakage between synthetic samples and
their real parents \cite{khosravi2024synthleak}, an instance of the
broader data-leakage problem in applied machine learning
\cite{kapoor2023leakage,apicella2025leakage}. Our parent-image fold
filter (Section~\ref{subsec:leak_free}) closes exactly this path, and
MaskCN avoids it by construction.

\subsection{Ensemble learning and stacked generalisation}
\label{subsec:rw_ensembles}

Combining several segmentation models reduces prediction variance and
often exceeds the best single base learner, though, as our results show
(Table~\ref{tab:stacking_results}), not always. Stacked ensembles have
been applied to concrete-crack and damage segmentation
\cite{li2022stackingconcrete,lee2023crackstacking}, and deep ensembles
are a standard route to accuracy and calibrated uncertainty
\cite{lakshminarayanan2017deepensembles}. Simple strategies such as
pixel-level majority voting and weighted probability averaging
\cite{thapa2025thesis} are easy to implement but carry two weaknesses
here: they treat base models as interchangeable, ignoring the
per-architecture calibration that follows from temperature scaling
\cite{guo2017calibration,wang2023calibratingsegmentation}, and they
admit a leakage mode whenever per-model weights are tuned on the same
split later used to report ensemble accuracy.

Cross-validated stacking, formalised by Wolpert
\cite{wolpert1992stacking}, eliminates both. The training set is
partitioned into $K$ folds, each base model is trained once per fold and
predicts on the held-out fold, and the resulting out-of-fold predictions
are each produced by a model that did not observe the corresponding
example, so the estimate is unbiased. A meta-learner is then fitted on
that out-of-fold matrix. Although stacking is ubiquitous in tabular
learning, its use for pixel-level semantic segmentation has been
limited. The present paper instantiates it at the pixel level, prepends
a per-architecture temperature calibration step so that base
probabilities are comparable, and pairs it with a parent-image filter
that keeps synthetic descendants of held-out images out of the
corresponding training folds. Calibration is a complementary literature:
Guo \textit{et al.}\ \cite{guo2017calibration} showed that modern deep
classifiers are systematically overconfident and that a single positive
temperature scalar fitted on held-out data suffices to recalibrate them.
We apply this architecture-by-architecture to the out-of-fold cube,
since uncalibrated probabilities let the most overconfident architecture
dominate the meta-learner regardless of its accuracy.

\subsection{Hard-negative mining}
\label{subsec:rw_hardneg}

Hard-negative mining \cite{shrivastava2016hardnegative} reduces false
positives by preferentially training on negatives the current model
misclassifies, and classically operates on an unlabelled real pool. We
substitute a controllable, infinitely extensible synthetic pool: a
defect-disabled preset of the upstream pipeline
\cite{thapa2026diffusion} generates visually confusable negatives on
demand, the publicly released baseline segmenter scores each one, and
the sampler weights it in proportion to how strongly it fools that
baseline. This combination of a controllable synthetic pool with
baseline-driven confusion weighting is, to our knowledge, novel in the
civil-infrastructure defect-segmentation literature
(Section~\ref{subsec:hard_negatives}).

Against this backdrop, two methodological moves address gaps the
ensembling and leakage literature leaves open: leak-free cross-validated
stacking with per-architecture calibration, instantiated at the pixel
level; and a parent-image fold filter that closes the silent leakage
path through synthetic descendants. The empirical contribution is an
honest account of which of these additions transfer to a small,
fuzzy-boundary defect and which do not.

\section{Methodology}
\label{sec:methodology}

\begin{figure*}[t]
\centering
\includegraphics[width=0.98\textwidth]{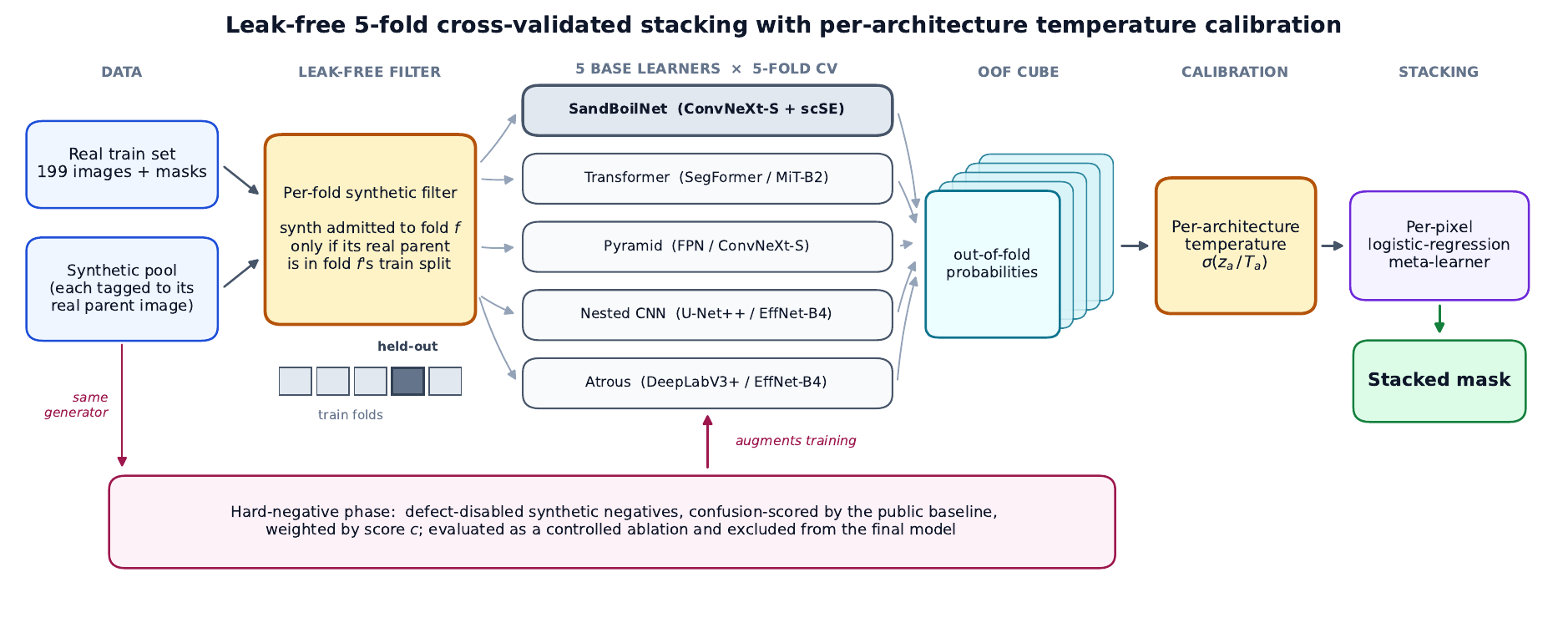}
\caption{The segmentation framework at a glance. Real and synthetic data pass
through the leak-free per-fold filter; five backbones are trained under
five-fold cross-validation; their out-of-fold probabilities are
recalibrated by one temperature scalar per architecture and combined by a
per-pixel meta-learner. The lower branch is the synthetic hard-negative
phase, evaluated as a controlled ablation and excluded from the final
model. The two steps drawn in amber are this paper's methodological
contributions; neither is an accuracy lever, since the filter removes an
upward bias rather than raising the score and calibration leaves test IoU
unchanged while making the weights comparable.}
\label{fig:pipeline_overview}
\end{figure*}

The segmentation framework comprises four coupled components: five
encoder--decoder backbones trained under five-fold
cross-validation, a parent-image filter that closes the synthetic
descendant leakage path, a per-architecture temperature calibration
step, and a per-pixel meta-learner that combines the calibrated
out-of-fold probabilities into a single stacked output. A
synthetic hard-negative training phase is added on top of the
principal training run, in which the publicly released baseline
segmenter scores each synthetic negative and the negative-phase
sampler weights each one in proportion to its confusion score.
Figure~\ref{fig:pipeline_overview} shows the framework as a whole;
Figures~\ref{fig:stacking_train} and~\ref{fig:stacking_test} expand its
training and inference halves.

\subsection{Dataset}
\label{subsec:dataset}

\begin{figure*}[t]
\centering
\includegraphics[width=0.92\textwidth]{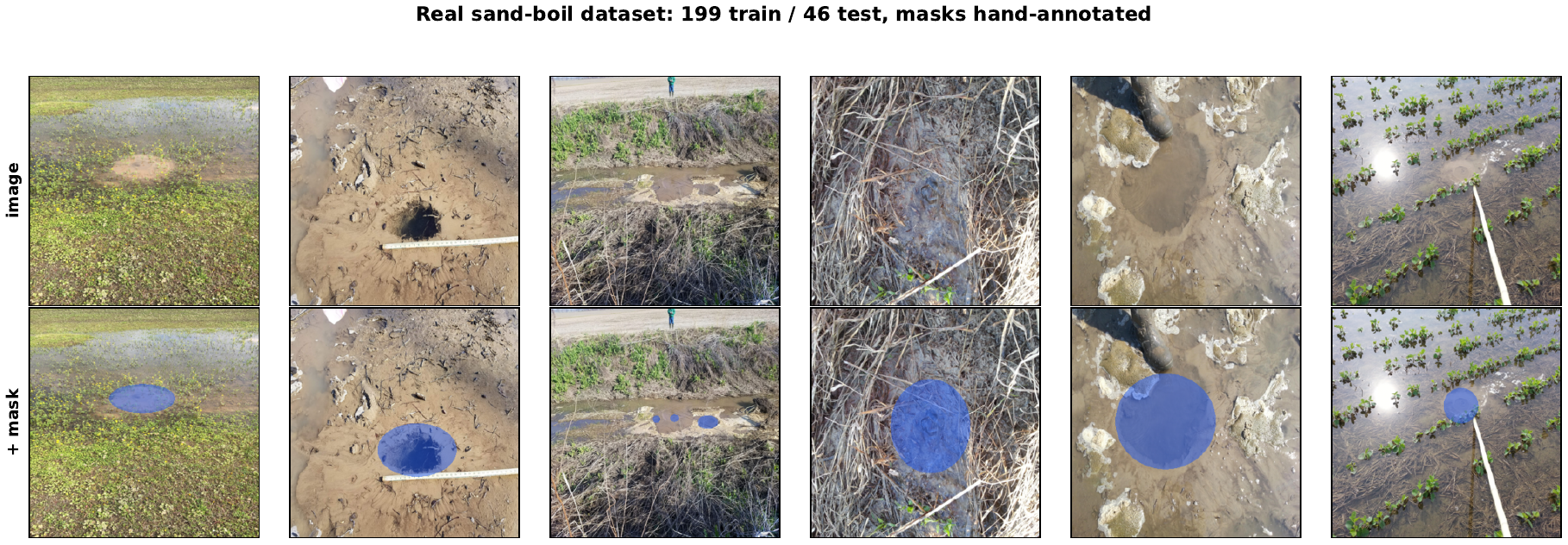}
\caption{Representative real sand boils (top) with their hand-annotated
ground-truth masks overlaid (bottom). The dataset spans a wide range of
boil sizes, water clarity, vegetation, and lighting; 199 images are used
for training and 46 are held out for test.}
\label{fig:dataset_montage}
\end{figure*}

The real sand-boil dataset is drawn from the U.S.\ Army Corps of
Engineers (USACE) levee-inspection archive
\cite{usace_levee_manual}; Figure~\ref{fig:dataset_montage} shows
representative images with their ground-truth masks. Field
inspectors traverse the crest and
adjacent areas of an earthen levee and photograph any
abnormalities they encounter. A domain-expert-curated subset of
approximately three hundred photographs was retained on the basis
of visibility and unambiguous appearance, then further filtered by
automatic intersection-over-union agreement between the manual
annotation and an initial Segment Anything proposal
\cite{kirillov2023sam}. The final curated set used in the present
study contains $N_{\mathrm{train}}{=}199$ training images and
$N_{\mathrm{test}}{=}46$ held-out test images at variable native
resolutions, each paired with a binary pixel-level mask. The
prose elsewhere refers to ``the training set'' and ``the held-out
test set'' to avoid restating these absolute numbers. For
downstream segmentation, images and masks are resampled to
$512 \times 512$ pixels.

\subsection{Upstream synthetic-data source}
\label{subsec:data_sources}

\begin{figure*}[t]
\centering
\includegraphics[width=0.92\textwidth]{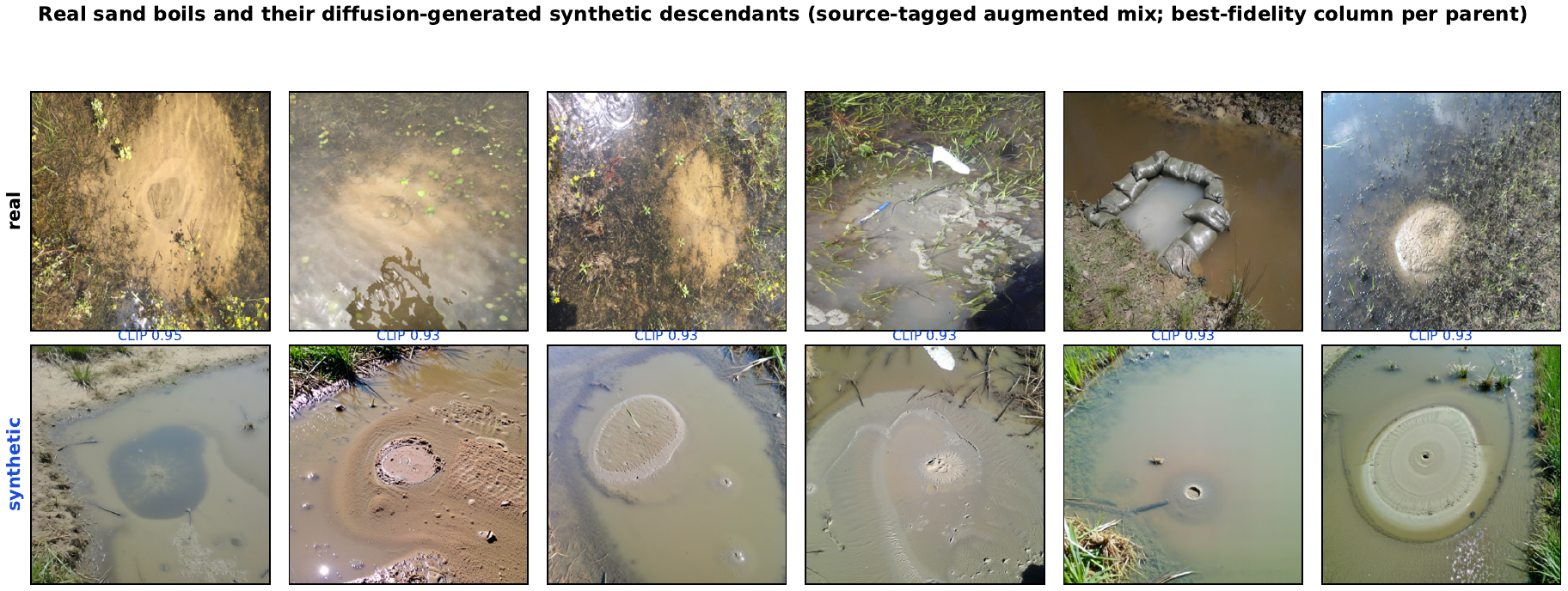}
\caption{Real sand boils (top) beside their best-fidelity synthetic descendants
(bottom), produced by the companion generation paper
\cite{thapa2026diffusion} and carried here in the source-tagged augmented
mix. Each bottom tile is the $V_4$-preset descendant scoring highest
under CLIP image similarity to the parent above it, and each carries a
\texttt{source\_image\_id} pointer to that parent, the field on which the
filter of Section~\ref{subsec:leak_free} keys.}
\label{fig:realsynth_montage}
\end{figure*}

The augmented training set and the synthetic hard-negative pool
consumed by the framework below are both produced by the
diffusion-based generation pipeline of our companion paper \cite{thapa2026diffusion};
Figure~\ref{fig:realsynth_montage} places representative synthetic
images beside their real parents. Our group has previously
released public synthetic levee-defect datasets in this line of
work through IEEE DataPort, covering sand boils
\cite{thapa2024dataport} and animal burrows \cite{thapa2025dataport2}.
We recap only what the segmentation framework needs to know.

The augmented training set comprises five synthetic variations of
every real training image, generated by a DreamBooth-fine-tuned
Stable Diffusion XL backbone with multi-ControlNet conditioning
under a soft-mask inpainting protocol that preserves the real
sand-boil region pixel-for-pixel and re-renders the surrounding
scene. Each synthetic record carries a
\texttt{source\_image\_id} pointer to its real parent (the
\emph{single} field the present paper requires from the upstream
pipeline) together with the prompt, seed, and generator preset.
Candidates are admitted only where all five cross-validation folds of
a held-out segmenter reach an intersection-over-union of $0.70$ against
the image's own mask, a label-fidelity filter applied uniformly across
presets; the survivors then pass a CLIP-similarity quality gate and are
sub-sampled by greedy farthest-point selection in CLIP feature space, so
the retained pool is diverse rather than merely accurate.
The synthetic ground-truth mask is produced by running
the synthetic image through the \textsc{Convex\_Hull\_Annotator}
pipeline of \cite{thapa2025thesis, thapa_chull_annotator}, which
thresholds the public SandBoilNet probability map and cleans the
result with connected-component analysis and convex-hull deburr. For
the quality-filtered pool of Section~\ref{subsec:res_ablation} the
labels are instead re-derived with the strongest segmenter available
here and morphological closing only, since a convex hull cannot
represent the concave rim of a boil.

The synthetic hard-negative pool is produced under a separate
production preset (denoted $V_{\mathrm{neg}}$) of the same
pipeline. $V_{\mathrm{neg}}$ disables the sand-boil DreamBooth
adapter at inference and drives the base SDXL backbone with a
category-specific prompt and a strong negative-prompt clause
suppressing residual sand-boil structure. Synthetic negatives
therefore share the visual style of the positive synthetic pool
(same SDXL backbone, same VAE, same camera-look) without
inheriting the sand-boil bias of the adapter.

\subsection{Mask-conditioned synthesis (MaskCN)}
\label{subsec:maskcn}

\begin{figure*}[t]
\centering
\includegraphics[width=0.98\textwidth]{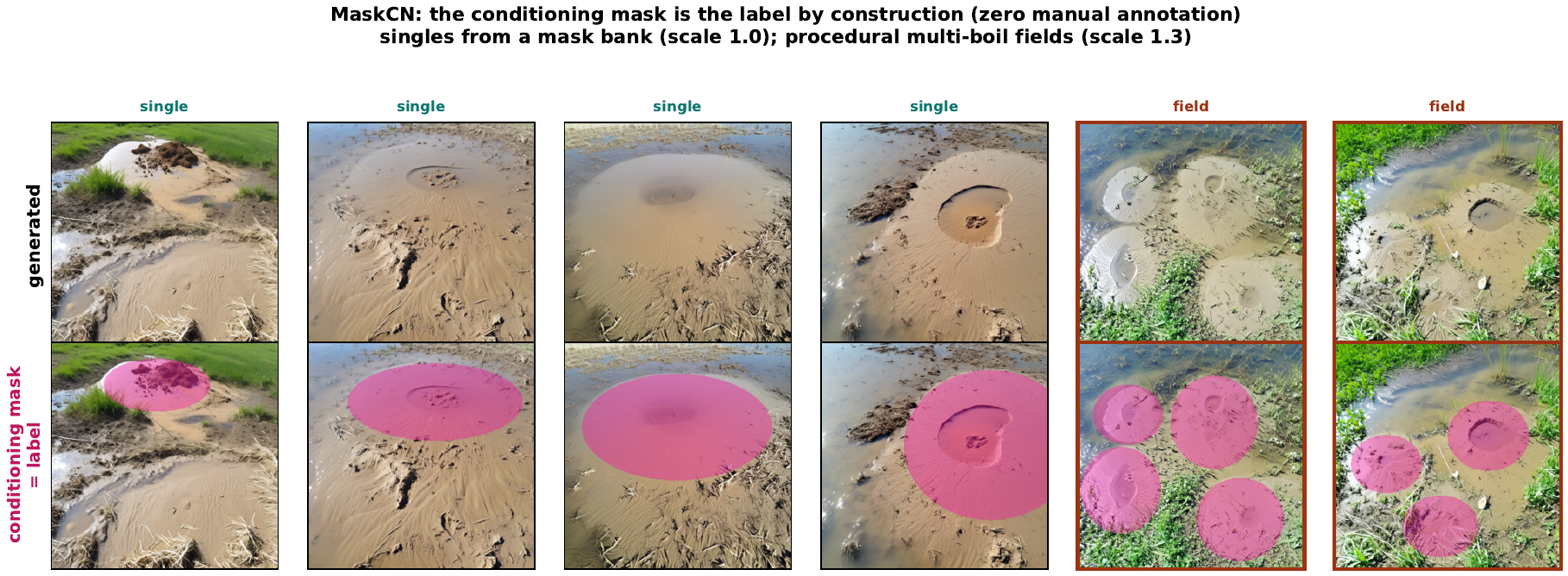}
\caption{Mask-conditioned synthesis (MaskCN). A mask-ControlNet
\cite{zhang2023controlnet} trained jointly with the sand-boil
DreamBooth-LoRA \cite{ruiz2023dreambooth,hu2022lora} renders a fresh sand
boil \emph{into} a target mask under a text-to-image preset, so the
conditioning mask (bottom row, magenta overlay) \emph{is} the
ground-truth label of the generated image (top row) by construction, at
zero annotation cost. Four single-boil tiles are conditioned on a mask
bank (ControlNet scale $1.0$); the two red-bordered tiles are procedural
multi-boil fields (scale $1.3$). MaskCN conditions on the mask alone. Its
value lies in zero-cost labelling and controlled diversity rather than in
raw accuracy (Table~\ref{tab:per_generator}).}
\label{fig:maskcn_showcase}
\end{figure*}

The augmented and hard-negative pools above both descend from a real
parent image and therefore carry the \texttt{source\_image\_id} that the
leak-free filter requires. We additionally introduce a complementary
\emph{mask-conditioned} synthesis route, \textsc{MaskCN}, whose records
have no real parent at all because the label is produced \emph{with} the
image rather than inferred from it
(Figure~\ref{fig:maskcn_showcase}). A second ControlNet
\cite{zhang2023controlnet}, a mask-ControlNet trained on
(mask, image) pairs together with the same sand-boil DreamBooth-LoRA
adapter \cite{ruiz2023dreambooth,hu2022lora}, is driven in
text-to-image mode and conditioned on a \emph{target} sand-boil mask. It
renders a new sand boil into that mask shape, so the conditioning mask
is the ground-truth annotation of the resulting image by construction,
requiring no \textsc{Convex\_Hull\_Annotator} pass and no manual
labelling. Conditioning masks come from two sources: a bank of $41$
real-derived mask silhouettes (single boils, ControlNet conditioning
scale $1.0$) and a procedural generator that scatters random ellipses to
form multi-boil \emph{fields} (conditioning scale $1.3$). All MaskCN
images are rendered at CFG $8.5$ with the DPM-Solver++ 2M Karras sampler
\cite{lu2022dpmsolver} and an fp16-fix VAE. Unlike the img2img augmented
pool, MaskCN conditions on the mask alone: it carries none of the Canny,
Depth, HED, Normal, denoising-strength, or IP-Adapter terms of the other
presets.

\begin{figure*}[t]
\centering
\includegraphics[width=0.64\textwidth]{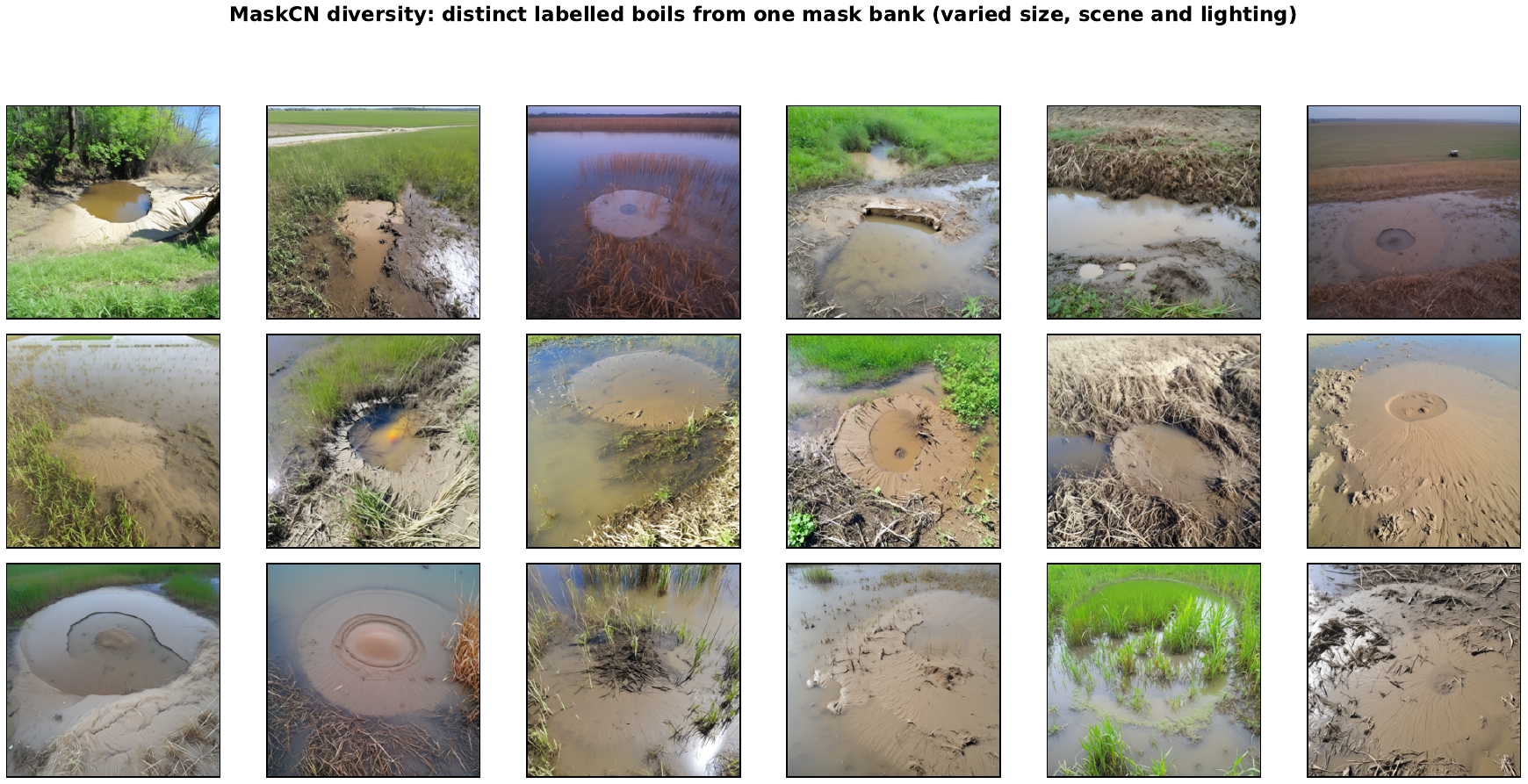}
\caption{MaskCN as a diversity engine. Eighteen distinct, fully labelled sand
boils synthesised from a single mask bank, ordered small-to-large in mask
coverage and varied in scene, framing, and lighting by per-source seed
variation. Because the conditioning mask supplies the label for every
tile, the pool extends to arbitrary size at no annotation cost.}
\label{fig:maskcn_diversity}
\end{figure*}

MaskCN matters to the segmentation framework for two reasons that are
independent of any accuracy claim. First, it is a \emph{zero-annotation
labelling} method: the most expensive ingredient in this domain is the
pixel-level mask, and MaskCN supplies a perfectly registered mask for
free with every image it generates. Second, it is a \emph{diversity}
engine: by sampling fresh boils into arbitrary mask shapes (including
procedurally generated multi-boil fields with no real-image
analogue), it produces labelled training material whose boil count,
size, placement, and scene context can be dialled independently of the
$199$-image real set (Figure~\ref{fig:maskcn_diversity}). Because MaskCN
records have no real parent, they sidestep the synthetic-descendant
leakage path of Section~\ref{subsec:leak_free}, with one caveat: the
conditioning masks themselves have sources, and a mask drawn from a
held-out image would leak that image's label geometry directly. The
evaluation pool therefore restricts the mask bank to training-set
sources (Appendix~\ref{app:maskcn}). We are explicit about scope: the
present paper introduces MaskCN as a labelling-and-diversity method.
Evaluated once as a direct augmentation source under the per-generator
protocol of Section~\ref{subsec:res_ablation}, its leak-safe pool is
competitive with the parent-derived pools
(Table~\ref{tab:per_generator}), though at that scale the differences
between pools are within run-to-run variation.

\subsection{Segmentation backbones}
\label{subsec:archs}

\begin{figure}[t]
\centering
\includegraphics[width=\columnwidth]{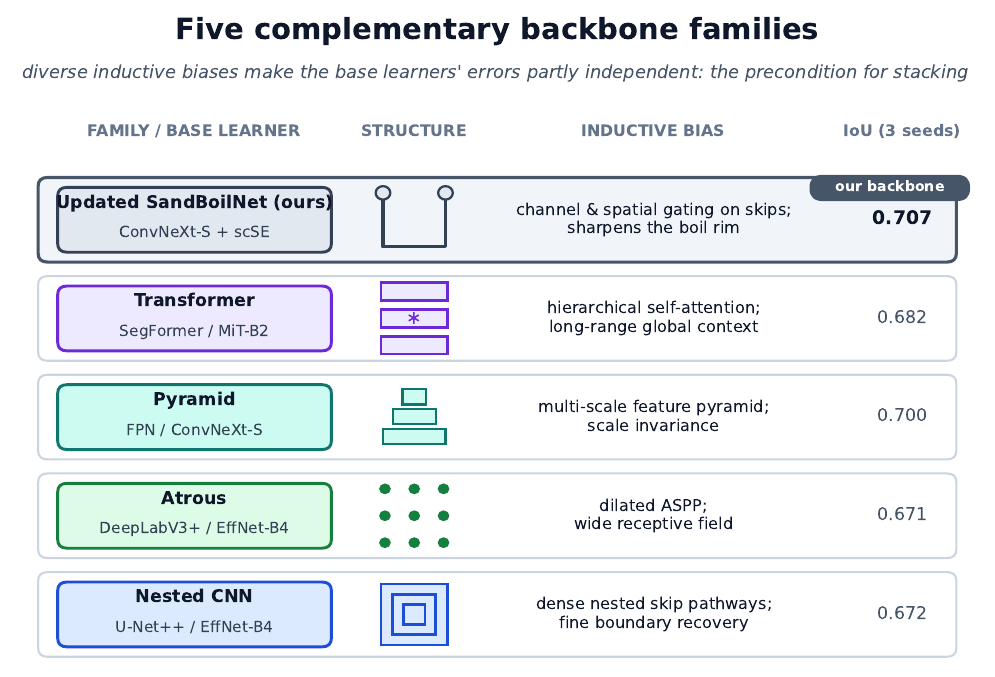}
\caption{The five backbone families retained as base learners, with the locked
encoder--decoder pairing and the inductive bias each contributes. The
families are chosen so their errors are partly independent, the condition
under which stacking improves on any single model, while a shared
$512\times512$ geometry keeps their predictions combinable. The right
column gives each champion's architecture-search IoU
(Table~\ref{tab:arch_search}). The highlighted row marks SandBoilNet as
the anchor of the ensemble rather than the single-split search winner.}
\label{fig:backbone_diversity}
\end{figure}

Rather than commit to one architecture, we run an architecture search
and then stack a small, deliberately diverse set of the strongest
models. The search covers eleven encoder--decoder networks that all
emit a single-channel probability map at the input's $512 \times 512$
resolution, drawn from five distinct architectural \emph{families}. The
diversity is the point: a stacking ensemble only helps to the extent
that its members fail in different places, so within each family we keep
only the best performer and discard the rest. Stacking two networks of
the same kind buys almost nothing, because they tend to make the same
mistakes on the same pixels. All encoders are ImageNet-pretrained, which
is decisive at this data scale, and each is trained with an
architecture-appropriate learning rate (lower for the transformers),
linear warmup, gradient clipping, and the heavy online augmentation of
Section~\ref{subsec:setup}. The five families
(Figure~\ref{fig:backbone_diversity}), and the model that
represents each in the final ensemble, are as follows.

\begin{enumerate}
\item \textbf{Attention U-Net (Updated SandBoilNet).} The
domain-specific SandBoilNet of Panta \textit{et al.}
\cite{panta2023sandboilnet} keeps a U-Net shape but enriches each skip
connection with attention. Its original Principal-Component-Analysis
attention is computationally heavy and numerically fragile; we replace
it with a lightweight spatial-and-channel squeeze-and-excitation (scSE)
gate \cite{roy2018scse} and swap the dated ResNet-50 encoder for a
ConvNeXt backbone \cite{liu2022convnext}. This Updated SandBoilNet is the
strongest single model once its folds are averaged (Table~\ref{tab:stacking_results})
and anchors the ensemble, though on the single-split search FPN edges it
(Table~\ref{tab:arch_search}).

\item \textbf{Transformer.} A hierarchical vision transformer
(SegFormer \cite{xie2021segformer}) replaces
convolutional locality with global self-attention, giving error modes
that differ sharply from the convolutional models and therefore
contribute the most independent signal to the stack.

\item \textbf{Nested convolutional U-Net.} U-Net++ \cite{zhou2018unetpp}
with an EfficientNet encoder \cite{tan2019efficientnet}, whose dense
nested skip pathways fuse features across decoder depths.

\item \textbf{Pyramid / multi-scale.} The best of a feature-pyramid
(FPN \cite{lin2017fpn}), pyramid-pooling (PSPNet \cite{zhao2017pspnet}),
or multi-scale-attention (MAnet \cite{fan2020manet}) decoder, all of
which aggregate context at several spatial scales.

\item \textbf{Atrous (dilated).} DeepLabV3+ \cite{chen2018deeplabv3plus},
whose atrous spatial-pyramid pooling widens the receptive field without
sacrificing output resolution.
\end{enumerate}

All five share a common input--output geometry, which is what makes their
out-of-fold probability maps directly stackable at the pixel level. Within
each family the search retains a single locked encoder--decoder
champion (U-Net++/EfficientNet-B4, SegFormer/MiT-B2, FPN/ConvNeXt-S,
DeepLabV3+/EfficientNet-B4, and the ConvNeXt-S$+$scSE SandBoilNet), and
these champions, with their held-out scores, are reported in
Table~\ref{tab:arch_search}; the original five U-Net
variants of our earlier work \cite{thapa2025thesis} are retained there
only as baselines. These five family champions are the stacking
members: the ensemble reported in Section~\ref{sec:results} is exactly
this set, one network per architecture family, with no substitutions.

\subsection{Texture-sensitive skip attention}
\label{subsec:feature_attn}

A sand boil is distinguished from the surrounding mud, water, and
vegetation chiefly by its fine sandy \emph{texture}, so we examine whether a
texture-sensitive skip-connection gate helps the champion backbone. The
scSE attention recalibrates decoder features with first-order (mean)
channel statistics and a single-map spatial gate
\cite{roy2018scse,hu2018senet}. We study a texture-sensitive variant,
\emph{Texture-SE}, that replaces the mean channel descriptor with a
second-order (standard-deviation) one, so a channel's texture energy rather
than its average activation informs its weight. Two further variants probe
whether additional machinery helps: a texture--contrast form (TC-SE) that
also replaces the spatial gate with a center--surround contrast term
$x-\mathrm{blur}(x)$, and a multi-scale dilated spatial form (Atrous-SE).
All add fewer than $0.2$\,M parameters and, unlike the PCA attention of the
original SandBoilNet, involve no eigendecomposition and are numerically
stable. The ablation of Section~\ref{subsec:res_ablation} finds that none
of these gates improves on the plain scSE skip attention on the present
split, so the champion retains scSE and the stacking ensemble is built from
the five architecture families rather than from attention variants of one
backbone.

\subsection{Cross-validated stacking}
\label{subsec:stacking}

\begin{figure}[t]
\centering
\includegraphics[width=\columnwidth]{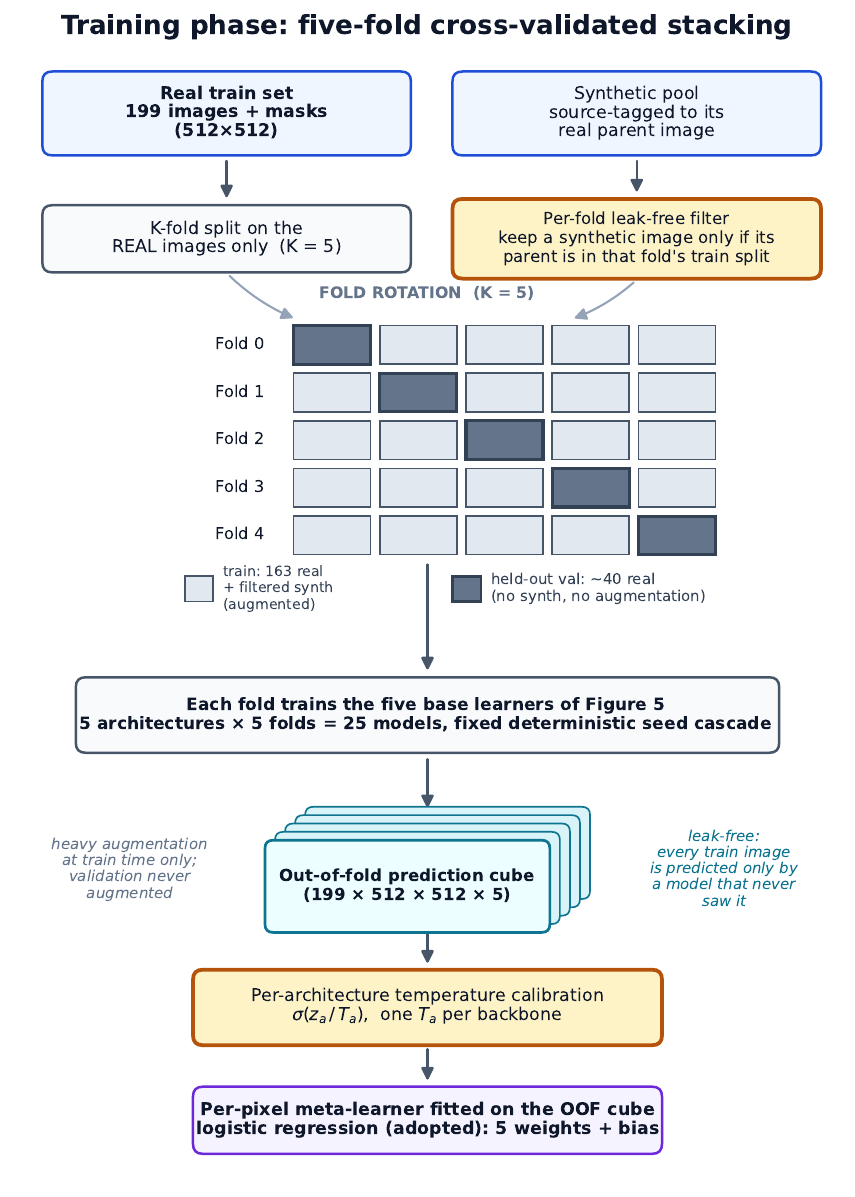}
\caption{Training phase. The real training set is split into five folds on the
real images only; the optional synthetic pool passes a per-fold
leak-free filter that keeps a synthetic image only when its
\texttt{source\_image\_id} parent lies in that fold's training split. In
the fold-rotation matrix the amber cell marks each fold's held-out slice,
so every real image is held out exactly once. Each base learner is
trained once per fold (25 models), and the held-out-fold predictions form
the out-of-fold cube that is calibrated and combined. Test-time inference
is shown in Figure~\ref{fig:stacking_test}.}
\label{fig:stacking_train}
\end{figure}

\begin{figure}[t]
\centering
\includegraphics[width=\columnwidth]{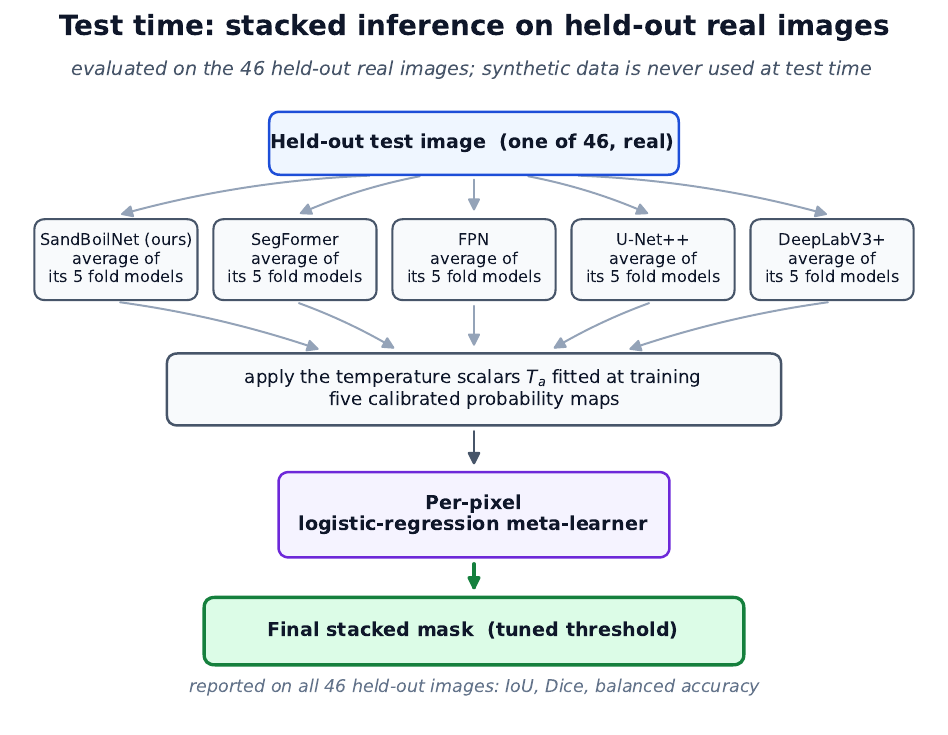}
\caption{Test-time inference. For each of the 46 held-out real test
images, every architecture averages its five fold models and applies its
fitted temperature scalar, yielding five calibrated probability maps.
The per-pixel meta-learner combines them into the final binary mask,
scored by IoU, Dice, and balanced accuracy. Synthetic data is never used
at test time.}
\label{fig:stacking_test}
\end{figure}

Figures~\ref{fig:stacking_train} and~\ref{fig:stacking_test} give the
training and inference halves of the pipeline; the remainder of this
section specifies each stage. The five backbones are trained under
five-fold cross-validation.
The real training set is partitioned into $K{=}5$ approximately
equal folds via a seeded shuffle. The partitioning seed is fixed
at $42$ and recorded in the artefact manifest, so the same fold
assignment is reproduced on every run. For each fold $f$ and each
architecture $a$ a model is trained on
the union of the other $K-1$ folds and predicts on the held-out
fold, producing one out-of-fold prediction per real training
image per architecture. The encoders start from ImageNet-pretrained
weights and the decoders from random initialisation; each
(architecture, fold) pair draws an independent seed from a fixed
deterministic cascade, $\mathrm{seed} = \mathrm{seed}_0 + 1000\,a + f$,
so the fold assignment and every model initialisation are
reproducible from the base seed, while the fold models of a given
architecture differ in both their training data and their
initialisation, which adds benign diversity to the out-of-fold
ensemble. One caveat is recorded for exactness: the augmentation
library draws its transform parameters from an internal generator
that this cascade does not govern, so bit-exact repetition of a
training run additionally depends on the pinned augmentation
library version. The
full out-of-fold collection is stored as a memory-mapped float-16
array of shape $(N_{\mathrm{train}}, H, W, A) = (199, 512, 512,
5)$ with a sidecar JSON manifest listing the image identifiers in
row order. Each (architecture, fold) write goes through the
manifest, so the procedure is crash-safe and resumable.

\paragraph{Fold-model health screen.} Ensemble quality is
sensitive to a failure mode that per-fold validation IoU alone
does not expose: a fold model occasionally converges to a solution
whose background probabilities floor well above zero (around
$0.08$ in our runs) instead of decaying toward it. The thresholded
masks, and therefore the validation IoU, can remain unremarkable,
but the inflated background mass distorts that architecture's
temperature fit and feeds the meta-learner a mis-scaled column.
Every (architecture, fold) model is therefore screened on its
out-of-fold predictions before any calibration or meta-learner
fitting: a model is accepted only if at least half of its
out-of-fold pixels receive a probability below $0.1$ (healthy fold
models assign well over $80\%$) and its held-out IoU reaches
$0.5$. A model that fails the screen is retrained with a fresh
seed drawn from the same deterministic cascade, up to four
attempts, keeping the healthiest attempt. The screen consults
training-side out-of-fold data only, so it introduces no test
leakage.

\paragraph{Fold count and nested stacking.} The out-of-fold
predictions used to fit the meta-learner come from \emph{single}-fold
models (trained on $K-1$ folds), whereas at test time each
architecture is the average of its $K$ fold models. To probe the
effect of this mismatch we ablate the fold count, training the stack
at $K{=}5$ and $K{=}10$ (the latter giving each fold model $90\%$ of
the data, so its out-of-fold prediction is closer to the
fold-averaged test model), and we additionally evaluate a
\emph{nested} cross-validated stack in which the meta-learner is
trained on fold-averaged predictions: for each outer fold an inner
$K{=}5$ cross-validation produces a five-model average on the
held-out outer fold, matching the test-time averaging. The three
variants are compared in Section~\ref{subsec:res_stacking}.

\subsubsection{Leak-free synthetic-data inclusion}
\label{subsec:leak_free}

\begin{figure}[t]
\centering
\includegraphics[width=\columnwidth]{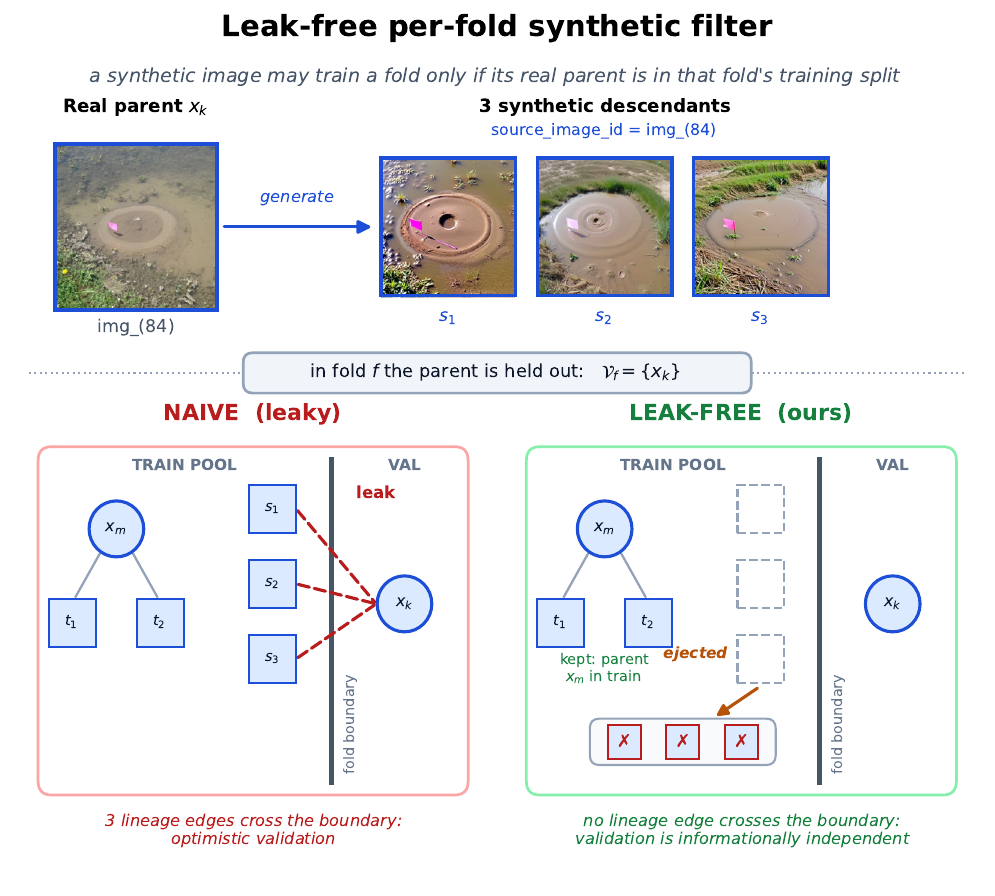}
\caption{The leak-free per-fold filter as a constraint on the fold boundary.
Top: a real parent and three of its synthetic descendants, linked by
\texttt{source\_image\_id}. Bottom: when fold $f$ holds out the parent
$x_k$, the naive pool (left) keeps its descendants $s_1$--$s_3$ in
training, so three lineage edges cross the fold boundary and validation is
optimistic. The filter (right) ejects exactly those descendants via
Eq.~\eqref{eq:leakfree}, while the family of a parent $x_m$ still in
training is retained.}
\label{fig:leakfree}
\end{figure}

The parent-tagged synthetic record structure makes the leak-free
filter trivial. For every fold $f$, the set of real identifiers
allocated to fold $f$'s validation split is collected, and any
synthetic image whose \texttt{source\_image\_id} falls in that
set is excluded from fold $f$'s training pool. No fold ever sees
a synthetic descendant of one of its own validation images. The
filter reduces to a one-line set intersection per fold:
\begin{equation}
\mathcal{T}^{\mathrm{aug}}_f =
\mathcal{D}^{\mathrm{aug}} \setminus
\big\{ s \in \mathcal{D}^{\mathrm{aug}} :
\mathrm{parent}(s) \in \mathcal{V}_f \big\} ,
\label{eq:leakfree}
\end{equation}
where $\mathcal{D}^{\mathrm{aug}}$ is the full augmented set,
$\mathcal{V}_f$ is the set of real validation identifiers for
fold $f$, and $\mathrm{parent}(s)$ is the \texttt{source\_image\_id}
of synthetic record $s$. The filter is $O(|\mathcal{D}^{\mathrm{aug}}|)$
and runs once at fold-construction time; Figure~\ref{fig:leakfree}
illustrates how it removes a held-out parent's synthetic descendants
from that fold's training pool.

\subsubsection{Per-architecture temperature calibration}

Different segmentation backbones produce probability outputs over
different confidence ranges. Naively stacking uncalibrated
probabilities allows the most overconfident architecture to
dominate the meta-learner output, regardless of whether it is the
most accurate. We therefore apply temperature scaling
\cite{guo2017calibration} to each architecture's out-of-fold
predictions: a single positive scalar $T_a$ is fitted to minimise
binary cross-entropy against the real ground-truth masks, and the
calibrated probability is
\begin{equation}
p^{\mathrm{cal}}_a = \sigma\!\big( \sigma^{-1}(p_a)\,/\,T_a \big) ,
\label{eq:tempscale}
\end{equation}
where $\sigma$ is the sigmoid function. The scalar is fitted by a
one-dimensional grid search over $26$ evenly spaced values of
$T_a \in [0.5, 3.0]$, evaluated on a fixed-seed subsample of one
million out-of-fold pixels, requiring no automatic
differentiation. A temperature $T_a > 1$ indicates that
the architecture was overconfident and was softened; $T_a < 1$
indicates that it was underconfident and was sharpened.

\subsubsection{Per-pixel meta-learner}

The meta-learner consumes the five calibrated probabilities at one
pixel and outputs a single probability for that pixel. The default
realisation is logistic regression \eqref{eq:metaforward}, parameterised
by a five-vector $\mathbf{w}$ of architecture weights and a scalar bias
$b$:
\begin{equation}
p^{\mathrm{stack}}(i,j) =
\sigma\!\left( \mathbf{w}^{\top} \mathbf{p}^{\mathrm{cal}}(i,j) + b \right) ,
\label{eq:metaforward}
\end{equation}
where $\mathbf{p}^{\mathrm{cal}}(i,j) =
(p^{\mathrm{cal}}_1(i,j), \ldots, p^{\mathrm{cal}}_5(i,j))$.
Training minimises $\ell_2$-regularised binary cross-entropy
($C{=}1$) on a fixed-seed subsample of two million pixels drawn
from the out-of-fold cube. A useful by-product of the linear form is
interpretability: each $w_a$ admits a direct reading as the
relative contribution of architecture $a$ to the final stacked
output. The choice of meta-learner is itself an ablation axis: we
compare logistic regression against seven alternatives: a
multi-layer perceptron, linear, RBF and polynomial SVMs, a random
forest, histogram gradient boosting, and XGBoost
\cite{chen2016xgboost}, all fitted on the same calibrated
out-of-fold cube with the exact configurations recorded in
Appendix~\ref{app:meta} and reported in
Table~\ref{tab:meta_compare}. We adopt logistic regression for its
directly interpretable weights.

Algorithm~\ref{alg:leakfree_stack} collects the stages above into the
procedure actually run. Two properties are visible directly from the
listing rather than argued in prose: the fold partition is drawn on the
real images alone (line~\ref{ln:partition}), and the parent-image filter
is applied \emph{inside} the fold loop, before any training
(line~\ref{ln:filter}), so no fold can train on a synthetic descendant of
one of its own validation images. Both the temperature scalars and the
meta-learner are fitted only on out-of-fold predictions
(lines~\ref{ln:temp} and~\ref{ln:meta}); the held-out test set is first
touched at line~\ref{ln:test}. The ensemble's accuracy is therefore
unbiased by construction rather than by assumption.

\begin{algorithm}[t]
\caption{Leak-free $K$-fold stacking with per-architecture calibration}
\label{alg:leakfree_stack}
\small
\begin{algorithmic}[1]
\Require real set $R$; synthetic pool $S$ (each image tagged with its real
parent); architectures $A$; folds $K$
\Statex\hspace{-1.2em}\textbf{Train} ($|A|{\times}K$ models, out-of-fold cube):
\For{each fold $f$ and architecture $a$} \label{ln:partition}
  \State $S_f \gets \{\,s \in S : \mathrm{parent}(s) \notin f\,\}$
         \Comment{leak-free filter, Eq.~\eqref{eq:leakfree}} \label{ln:filter}
  \State train $M_{a,f}$ on $(R \setminus f) \cup S_f$; store its fold-$f$
         predictions
\EndFor
\Statex\hspace{-1.2em}\textbf{Calibrate + combine} (out-of-fold only):
\State per architecture, fit temperature $T_a$ and rescale
       \Comment{Eq.~\eqref{eq:tempscale}} \label{ln:temp}
\State fit per-pixel meta-learner on the calibrated cube; pick threshold
       $\tau$ \label{ln:meta}
\Statex\hspace{-1.2em}\textbf{Predict}: for test image $x$, average each
architecture's $K$ folds, rescale by $T_a$, combine, threshold at $\tau$.
\label{ln:test}
\end{algorithmic}
\end{algorithm}

\subsubsection{Test-time inference}

For each held-out test image, the $K{=}5$ fold checkpoints of
each architecture produce $K$ predictions; these are averaged
within architecture (a form of bagging that reduces variance),
then passed through the per-architecture temperature and into the
meta-learner, which produces the final stacked segmentation.

\subsection{Hard-negative phase}
\label{subsec:hard_negatives}

\begin{figure}[t]
\centering
\includegraphics[width=\columnwidth]{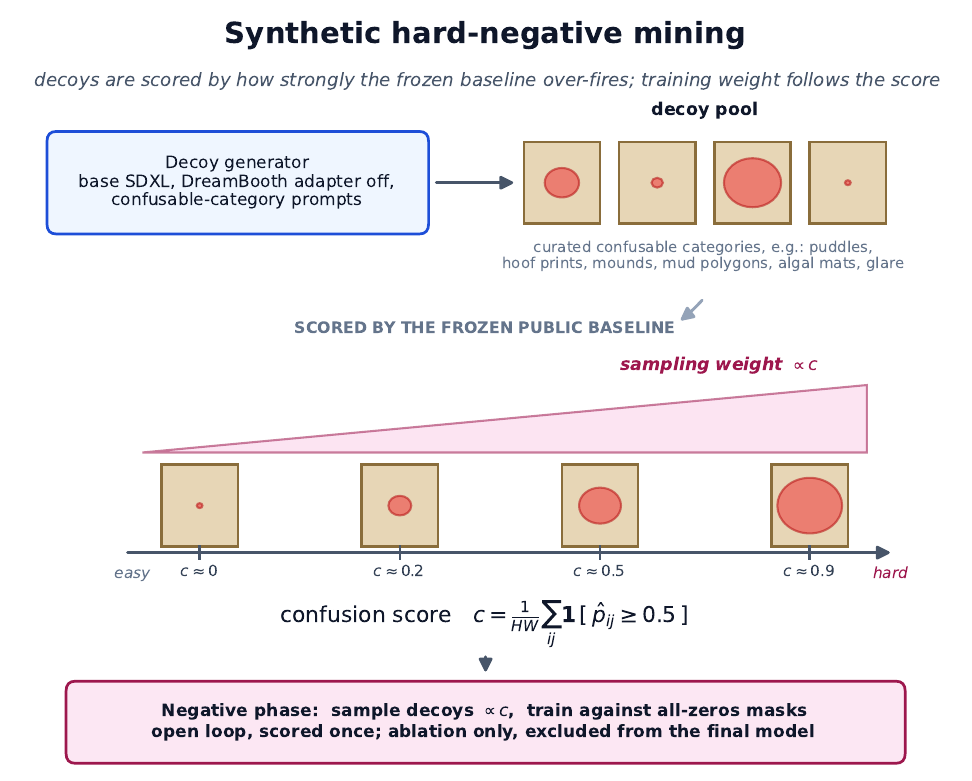}
\caption{Synthetic hard-negative mining as a curriculum. An adapter-off
generator renders confusable non-boil decoys; the frozen public
baseline's over-firing places each decoy on the confusion-score axis
(red blob: the baseline's false-positive response; tiles illustrative),
and the negative phase samples decoys with weight $\propto c$
(Eq.~\eqref{eq:confusion_score}) against all-zeros masks. The loop is
open (each decoy is scored once), and the phase is evaluated as a
controlled ablation and excluded from the final model.}
\label{fig:hardneg}
\end{figure}

We adopt the negative-image training protocol of Panta
\textit{et al.}\ \cite{panta2023sandboilnet}. After the principal
training run on annotated sand-boil images, each segmentation
model is exposed to defect-free levee imagery (grassy crowns,
undisturbed embankments, clean water channels) through a short
second phase. The negative phase teaches the network to reject
visually similar but non-defect regions and reduces the
false-positive rate at inference. Approximately eighty real
defect-free levee photographs are drawn from the same USACE
inspection archive as the positive set, ensuring that lighting
and camera profile are matched.

\paragraph{Synthetic hard-negative atlas.} On top of the real
defect-free pool we generate a diffusion-synthesised pool of
\emph{visually confusable} hard negatives that target classes the
public SandBoilNet baseline is known to over-fire on
(Figure~\ref{fig:hardneg}). The
category list is curated from inspector-side reports of frequent
false positives and from a prior misclassification audit:
mud puddles with sediment rings, hoof-print puddles, tractor and
ATV rut puddles, mole or gopher mounds, anthills, cracked
dried-mud polygons, half-buried rocks and cobbles, algal mats,
drainage-tile outlets, subsidence puddles, sheet-erosion sediment
fans, catfish and turtle wallows, wet manure patties, reflective
standing-water glare, surfacing tree roots, and algal bloom
rosettes. For each category the upstream pipeline instantiates a
hard-negative prompt atlas of approximately fifty contextual
prompts using the same combinatorial-template machinery as the
principal prompt atlas, with category-specific descriptors of the
object combined with the shared field-context axes (geographic
setting, lighting, weather, camera medium, composition). Each
prompt carries a category-label tag and a negative-prompt extras
clause (``no sand boil, no sand mound, no domed sand cone, no
water erupting from the ground, $\ldots$'') that is appended
verbatim to the diffusion negative-prompt slot at inference time.

\paragraph{Generation preset.}\label{par:vneg} Negatives are
rendered through a production preset of the upstream pipeline
\cite{thapa2026diffusion}, denoted $V_{\mathrm{neg}}$, that reuses the prompt-driven scaffold
(weak Canny only, no HED, no Normal, no IP-Adapter, no soft-mask,
near-text2img CFG, high strength) but additionally sets the
DreamBooth-LoRA adapter weight to $0$ at inference. Disabling
the LoRA is essential here: the adapter has been fine-tuned to
\emph{produce} sand-boil domes from the curated reference set,
and leaving it at its default weight $1.0$ would re-introduce a
domed-mound bias into prompts that explicitly call for a puddle
or a mole hill. With the adapter weight at $0$, the base SDXL
backbone follows the textual prompt, and the negative-prompt
clause prunes residual sand-boil structure from the denoising
trajectory. Synthetic negatives are generated at $1024 \times
1024$ and share the visual style of the positive synthetic pool
(same SDXL backbone, same VAE, same camera-look as the LoRA's
reference set), avoiding the style-mismatch problem that mixing
a heterogeneous external image pool would introduce.

\paragraph{Confusion-score mining.} Synthetic hard negatives are
not equally useful: an image that the segmenter already
classifies correctly contributes little, whereas an image the
baseline \emph{misclassifies} as a sand boil is the most
informative supervision for the negative phase. We therefore run the public
SandBoilNet checkpoint of \cite{panta2023sandboilnet} over every
synthetic negative and write a per-image confusion score
\begin{equation}
c = \tfrac{1}{HW} \sum_{i,j} \mathbb{1}\!\left[\hat{p}_{ij} \geq 0.5\right] ,
\label{eq:confusion_score}
\end{equation}
i.e., the fraction of pixels the baseline labels positive. The
negative phase consumes the union of the real defect-free pool
and the synthetic hard-negative pool, with synthetic samples
ranked by $c$ and sampled with weight proportional to $c$ so that
the training loop focuses on the negatives the baseline is most
fooled by. This mechanism is a lightweight form of hard-negative
mining \cite{shrivastava2016hardnegative} adapted to a
synthesised pool: instead of mining the unlabelled real pool for
high-loss examples (the classical formulation), we mine a
controllable, infinitely extensible synthetic pool. Negatives are
trained against an all-zeros target mask, leaving the combined
BCE plus Dice loss well-defined. The hyperparameters of the
negative phase match the main phase except for a fixed learning
rate of $1\mathrm{e}{-4}$ and a cap of $30$ epochs.

\subsection{Loss functions and metrics}
\label{subsec:metrics}

Segmentation training minimises the combined binary cross-entropy
plus Dice loss
\begin{equation}
\mathcal{L} = w_{\mathrm{bce}}\, \mathcal{L}_{\mathrm{bce}}
            + w_{\mathrm{dice}}\, \mathcal{L}_{\mathrm{dice}} ,
\label{eq:bce_dice}
\end{equation}
with $w_{\mathrm{bce}} = w_{\mathrm{dice}} = 0.5$. The BCE term
encourages correct per-pixel classification; the Dice term
encourages overall region overlap, which compensates for the
heavy class imbalance (sand-boil pixels constitute approximately
$5\%$ of the image).

Evaluation reports five measures: intersection-over-union (IoU) and
Dice coefficient (Eq.~\eqref{eq:iou_dice}), mean IoU (mIoU), binary
accuracy, and balanced accuracy (Eq.~\eqref{eq:ba}). IoU and Dice both
measure region overlap; balanced accuracy averages sensitivity and
specificity and remains informative under class imbalance. The
hard-negative analysis additionally reports recall (sensitivity,
$TP/(TP+FN)$), one of the two components of \eqref{eq:ba}, together with
the false-positive rate ($FP/(FP+TN)$), the complement of the specificity
term in \eqref{eq:ba}.
\begin{equation}
\mathrm{IoU} = \frac{|P \cap G|}{|P \cup G|},\qquad
\mathrm{Dice} = \frac{2\,|P \cap G|}{|P| + |G|},
\label{eq:iou_dice}
\end{equation}
\begin{equation}
\mathrm{BA} = \tfrac{1}{2}\!\left(\frac{TP}{TP + FN} + \frac{TN}{TN + FP}\right).
\label{eq:ba}
\end{equation}

\paragraph{Boundary tolerance and uncertainty.} Sand-boil rims are
intrinsically fuzzy: the saturated dome fades into the surrounding mud
over several pixels, so the exact rim placement of the ground-truth
annotation is itself uncertain to within a few pixels. A strict IoU
penalises any pixel that disagrees, conflating genuine misses with
sub-annotation-noise boundary jitter. We therefore additionally report a
\emph{boundary-tolerant} IoU that counts a predicted boundary pixel as
correct if a ground-truth boundary pixel lies within a tolerance band of
$t$ pixels (and conversely), reporting $t{=}0$ (the strict micro-average)
alongside $t{=}2$\,px. To quantify the sampling uncertainty that the
small held-out set induces, we attach a non-parametric bootstrap
$95\%$ confidence interval to the champion's test IoU, resampling the
held-out images with replacement; these two diagnostics are reported in
Section~\ref{subsec:res_boundary}.

The reproducibility discipline that ties every reported number to
its fold partition, seed cascade, and parent-image filter is
detailed with the experimental setup in
Section~\ref{subsec:reproducibility_exp}.

\section{Experiments}
\label{sec:experiments}

The experimental protocol is organised into three phases: a
single-architecture baseline phase that establishes a floor for
each backbone in isolation, a stacking-matrix phase that compares
single-architecture, mean-ensemble, and cross-validated stacked
ensembles under real-only and real-plus-synthetic training, and a
hard-negative ablation phase that isolates the contribution of the
confusion-score-mined negative pool.

\subsection{Hardware and software}
\label{subsec:setup}

All experiments run on a SLURM-scheduled cluster of four NVIDIA
H100 NVL GPUs ($95$~GB HBM each). The experiment matrix (the
five-fold, five-architecture cross-validation runs together with the
ablation and baseline jobs) is dispatched across the four GPUs with
\texttt{sbatch}: the runs are mutually independent, so they are queued
together and execute in parallel, and the full matrix completes with
roughly four-way throughput. Each job requests a single GPU, trains its
models to completion and exits, streaming logs to disk; the jobs are
task-parallel and need no cross-GPU communication, so the matrix scales
with the number of available devices rather than requiring distributed
training. Per-job wall-clock times are reported in
Appendix~\ref{app:cost}.

Segmentation training and the cross-validated stacking runs use
PyTorch \cite{paszke2019pytorch} with automatic mixed-precision (AMP)
autocasting and pinned-memory, multi-worker data loading, the
Segmentation Models PyTorch library
\cite{iakubovskii2019smp} for the encoder--decoder backbones and
pretrained encoders, and Albumentations \cite{buslaev2020albumentations}
for the joint image--mask augmentation pipeline. The meta-learners
(logistic regression and the alternatives compared in
Table~\ref{tab:meta_compare}) and the per-architecture temperature
calibration run on CPU with scikit-learn \cite{pedregosa2011scikit}
and complete in under a minute.

\subsection{Segmentation training protocol}
\label{subsec:seg_protocol}

Each segmentation experiment uses the combined BCE plus Dice loss
\eqref{eq:bce_dice} with the AdamW optimiser and a batch size of
$8$. Encoders are ImageNet-pretrained; decoders are randomly
initialised. The learning rate is set per architecture and follows a
five-epoch linear warmup into cosine annealing: $1\mathrm{e}{-4}$
for the CNN baselines, $8\mathrm{e}{-5}$ for ConvNeXt encoders, and
$6\mathrm{e}{-5}$ for the transformer (weight decay $1\mathrm{e}{-4}$,
raised to $1\mathrm{e}{-2}$ for the transformer); the per-architecture
settings are tabulated in Appendix~\ref{app:training_config}. Training
caps at
$120$--$160$ epochs depending on the family, with early stopping on
validation IoU (patience $30$); the best-IoU checkpoint is restored
for testing. Spatial augmentations from a twenty-nine-transform
Albumentations \cite{buslaev2020albumentations} pipeline
(Figure~\ref{fig:aug_showcase}, Appendix~\ref{app:aug_pipeline}) are applied jointly to image and
mask; intensity and noise transforms are applied to the image alone.
Validation uses identical preprocessing without augmentation.

\subsection{Stacking experiment matrix}
\label{subsec:stacking_matrix}

The stacking experiments compare three ensembling strategies (best
single architecture; the mean ensemble of the backbones, with the prior
weighted-average ensemble of \cite{thapa2025thesis} as a reference row;
and leak-free cross-validated stacking) under two training-data
conditions: real only, and real plus synthetic under the leak-free filter
of Section~\ref{subsec:leak_free}. The held-out prediction of each
architecture is the average of its five fold models, so every reported
figure is a fold-averaged central estimate on the held-out real test set,
scored by IoU, Dice, mean IoU, binary and balanced accuracy, and macro-F1.
The headline comparison (Table~\ref{tab:stacking_results}) reports the
real-only condition; the effect of synthetic augmentation on the strong
ensemble is analysed in Section~\ref{subsec:res_ablation}. The choice of
meta-learner \emph{within} the cross-validated-stacking strategy (logistic
regression, an MLP, linear/RBF/polynomial SVMs, random forest, histogram
gradient boosting, and XGBoost) is a separate axis explored in
Table~\ref{tab:meta_compare}, holding the data and folds fixed.

\subsection{Hard-negative ablation}
\label{subsec:hardneg_ablation}

The hard-negative phase is evaluated through a two-condition
ablation. The \emph{baseline} condition is the principal training
run alone: Updated SandBoilNet trained on the real training set
plus the augmented synthetic pool under the leak-free filter of
Section~\ref{subsec:leak_free}, with no negative phase. The
\emph{negative-phase} condition applies, on top of that identical
run, a short additional phase trained on the synthetic
hard-negative pool of Section~\ref{subsec:hard_negatives}, with
samples weighted by their confusion score $c$ of
\eqref{eq:confusion_score} (learning rate $1\mathrm{e}{-4}$, cap
of $30$ epochs). The contribution of the synthetic negative pool
is read as the difference in false-positive rate (and in balanced
accuracy) between the two conditions on the held-out real test
set.

\subsection{Per-architecture meta-learner weight analysis}
\label{subsec:weights_analysis}

We report the meta-learner weights normalised so that
$\sum_a |w_a| = 1$, alongside the per-architecture calibration
temperatures $T_a$ of \eqref{eq:tempscale}. The two are read together in
Section~\ref{subsec:res_weights}.

\subsection{Reproducibility}
\label{subsec:reproducibility_exp}

Every numerical entry reported in Section~\ref{sec:results} is the
deterministic product of three identifiers: the fold partition
seed (fixed at $42$), the architecture and initialisation seed,
and the \texttt{source\_image\_id} list of every synthetic record
admitted to the training pool. Each segmentation training run
records the architecture name, fold index, initialisation seed,
optimiser hyperparameters, and a hash of its preprocessing
pipeline. The upstream generation pipeline of the companion paper
\cite{thapa2026diffusion} emits a manifest of every adapter, prompt-bank revision, and
seed cascade used to produce the augmented and hard-negative
imagery consumed here. Together these identifiers allow any
reported number to be re-derived from the released code and
configuration files without manual reconfiguration.

\section{Results and Analysis}
\label{sec:results}

We report the outcomes of the experimental phases of
Section~\ref{sec:experiments}: the architecture search, the stacking
matrix, the hard-negative ablation, an uncertainty analysis of the
headline result, and the per-architecture meta-learner weights. The full
combiner-by-design grid is given in Appendix~\ref{app:meta}.

\subsection{Architecture search}
\label{subsec:res_arch}

Before stacking, we select the base learners by an architecture search
across the five encoder--decoder families of
Section~\ref{subsec:archs}, all trained under the heavy augmentation and
per-architecture schedule of Section~\ref{subsec:setup}, keeping the best
performer in each family since two networks of the same kind contribute
little independent signal to the stack. Table~\ref{tab:arch_search} reports the resulting family
champions on the held-out test set, with the original SandBoilNet shown
for reference.

\begin{table*}[t]
\caption{Architecture search: per-family champions retained as stacking base
learners, as mean $\pm$ standard deviation of held-out test IoU over
three seeds ($\{1,2,42\}$) on a single 85/15 split. ``Updated
SandBoilNet'' is SandBoilNet modernised with a ConvNeXt-S backbone and
scSE skip attention. Under the fold-averaged stacking protocol the
strongest member is instead SegFormer
(Table~\ref{tab:stacking_results}), the two regimes training on different
amounts of data. The reference row is the published checkpoint
re-evaluated on this paper's $46$-image test set.}
\label{tab:arch_search}
\centering
\small
\begin{tabular}{@{}llc@{}}
\toprule
Family & Base learner & Test IoU (mean $\pm$ sd) \\
\midrule
Attention U-Net (Updated SandBoilNet) & ConvNeXt-S $+$ scSE & $\mathbf{0.707 \pm 0.003}$ \\
Pyramid / multi-scale & FPN / ConvNeXt-S & $0.700 \pm 0.012$ \\
Attention U-Net, no gate & ConvNeXt-S & $0.697 \pm 0.009$ \\
Transformer & SegFormer / MiT-B2 & $0.682 \pm 0.020$ \\
Nested CNN & U-Net++ / EfficientNet-B4 & $0.672 \pm 0.014$ \\
Atrous & DeepLabV3+ / EfficientNet-B4 & $0.671 \pm 0.021$ \\
\midrule
\textit{reference} & SandBoilNet (ResNet50V2 $+$ PCA), our run & $0.608$ \\
\bottomrule
\end{tabular}
\end{table*}

The Updated SandBoilNet leads at $0.707$ IoU, $0.099$ above the published
original on the same split and a relative gain of $16\%$. Its margin over
the nearest families is far smaller ($0.700$ for the pyramid variant,
$0.697$ for the gateless baseline), so the improvement follows mainly from
the modernised encoder rather than the attention gate, and it is the most
reproducible entry at $\pm 0.003$ over three seeds. That the same
configuration also leads a disjoint defect dataset
(Section~\ref{subsec:res_deepcrack}) is the stronger evidence that the
ranking is not an artefact of this split.

\subsection{Cross-validated stacking results}
\label{subsec:res_stacking}

Table~\ref{tab:stacking_results} reports the headline comparison: the
best single architecture, the mean ensemble, and leak-free
cross-validated stacking with the adopted logistic-regression
meta-learner, on real-only training at the OOF-tuned operating point. The
five architectures of Table~\ref{tab:arch_search} stack to $0.681$ IoU
against $0.694$ for the strongest fold-averaged member, so the stack does
not recover the best single model; the mean ensemble reaches the same
$0.681$, leaving the fitted meta-learner with no advantage over uniform
weighting. The effect
of synthetic training data is examined in
Section~\ref{subsec:res_ablation} and the choice of meta-learner in
Table~\ref{tab:meta_compare}.

\begin{table*}[t]
\caption{Headline comparison on the held-out real test set at the leak-free
out-of-fold operating point; each architecture's prediction averages its
five fold models. Neither the mean ensemble nor the calibrated stack
reaches the best single model, both landing $0.013$ IoU below it, and the
stack is indistinguishable from the simple mean. In this fold-averaged
regime the strongest member is Segformer, whereas the Updated
SandBoilNet leads when trained once on the full training split
(Table~\ref{tab:arch_search}); the two regimes need not agree.
$^{\dagger}$As published by the original authors on their own split; our
like-for-like re-evaluation appears in Table~\ref{tab:arch_search}.}
\label{tab:stacking_results}
\centering
\small
\begin{tabular}{@{}lcccccc@{}}
\toprule
Method (real-only) & IoU & Dice & mIoU & Bin.\,Acc. & Bal.\,Acc. & Macro-F1 \\
\midrule
Panta et al.\ 2023$^{\dagger}$ \cite{panta2023sandboilnet} & 0.574 & n/a & n/a & n/a & 0.855 & 0.731 \\
Prior weighted-avg ensemble \cite{thapa2025thesis} & 0.547 & 0.671 & 0.697 & 0.871 & 0.850 & n/a \\
\midrule
Best single: Segformer (MiT-B2) & \textbf{0.694} & 0.801 & 0.827 & 0.966 & \textbf{0.913} & 0.890 \\
SandBoilNet (ConvNeXt-S$+$scSE) & 0.676 & 0.793 & 0.818 & 0.966 & 0.909 & 0.886 \\
Mean ensemble (five members) & 0.681 & 0.791 & 0.821 & 0.967 & 0.906 & 0.885 \\
\textbf{CV stacking (LR meta), ours} & 0.681 & 0.790 & 0.821 & \textbf{0.967} & 0.895 & 0.885 \\
\bottomrule
\end{tabular}
\end{table*}

\begin{figure}[t]
\centering
\includegraphics[width=0.86\columnwidth]{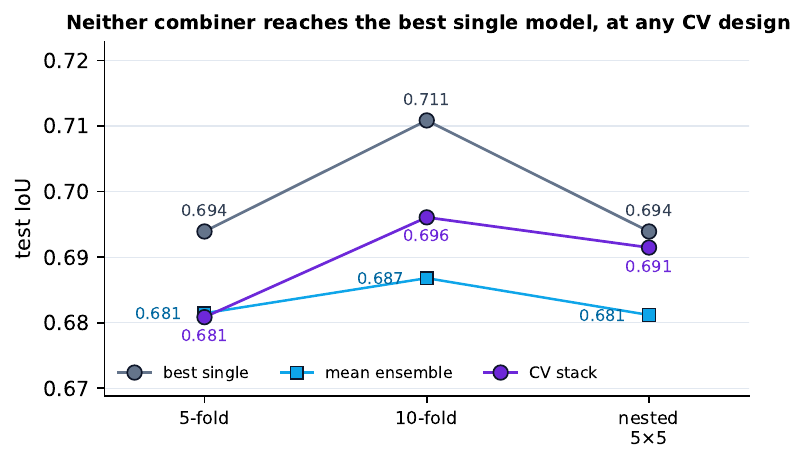}
\caption{Ensembling strategies against the best single model at each
cross-validation design (values in Table~\ref{tab:cv_variants}). No
combiner reaches the best single model at any design, and raising the fold
count lifts both together without closing the gap. The nested design comes
closest, its meta-learner being the only one fitted on the fold-averaged
predictions it meets at test time.}
\label{fig:cv_ladder}
\end{figure}

\begin{figure}[t]
\centering
\includegraphics[width=\columnwidth]{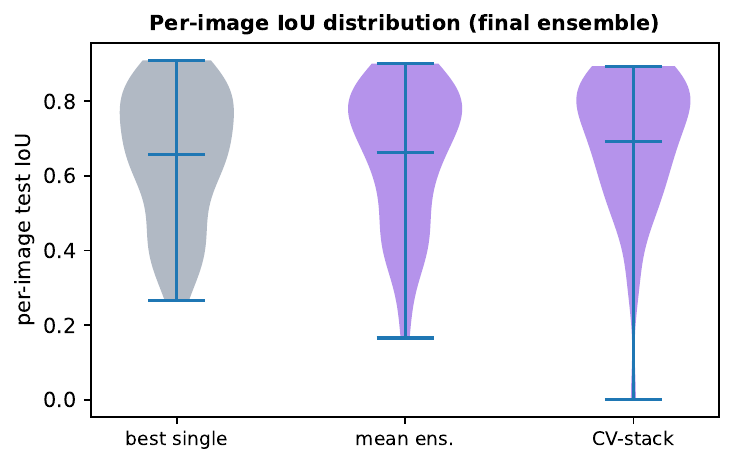}
\caption{Per-image IoU distributions for the best single base learner, the
mean ensemble, and the CV-stacked meta-learner of the final ensemble.
Stacking tightens the spread relative to the weaker single backbones, but
its mean IoU remains below the best single model
(Table~\ref{tab:stacking_results}): it trades a little worst-case variance
for a small loss in typical accuracy.}
\label{fig:iou_violin}
\end{figure}

\begin{figure}[!htbp]
\centering
\includegraphics[width=\columnwidth]{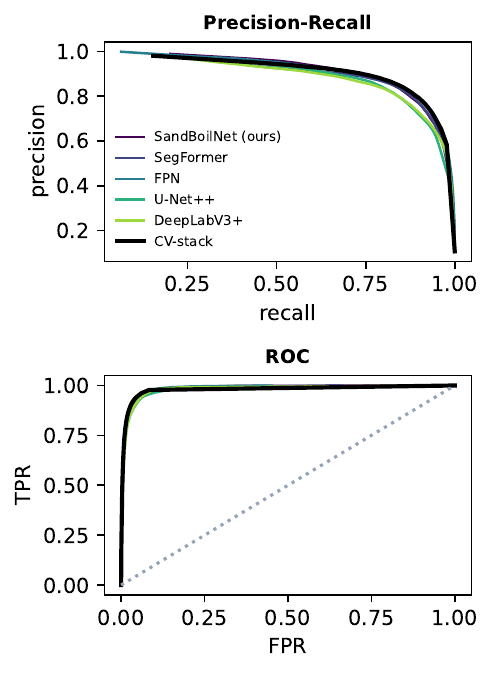}
\caption{Precision--recall (top) and ROC (bottom) curves across the five base
learners and the stacked ensemble. The stacked operating curve tracks the
best base learners across the whole sweep, so its behaviour does not rest
on a single threshold choice. The curves lie close together because the
five learners are the family champions of
Section~\ref{subsec:res_arch}: average precision spans only
$0.835$--$0.876$ and ROC AUC $0.967$--$0.981$, with the stack among the
strongest members rather than above them.}
\label{fig:proc}
\end{figure}

The comparison isolates the ensembling contribution against two
references. The prior weighted-average ensemble of \cite{thapa2025thesis}
(the practice that admits the mild leakage of
Section~\ref{subsec:rw_ensembles}) and the originally published
SandBoilNet are shown for context; the operative comparison is the best
single architecture versus the mean ensemble and the calibrated
cross-validated stacking of Section~\ref{subsec:stacking}, all on the same
five fold-averaged base learners. Under that comparison ensembling does
not pay: both combiners sit $0.013$ IoU below the strongest member, and
the fitted stack matches the unweighted mean to three decimal places,
which is what one expects when the members' errors are largely shared
rather than independent (Section~\ref{subsec:disc_stacking}).
Figure~\ref{fig:cv_ladder} plots these systems against the best-single
baseline across cross-validation designs, and Figure~\ref{fig:iou_violin} the
corresponding per-image IoU distributions, which show where the shortfall
lies: the stack compresses the lower tail slightly but does not lift the
median case, so it trades a little worst-case robustness for a small loss
in typical accuracy. Figure~\ref{fig:proc} gives the precision--recall and ROC curves for
the base learners and the stacked ensemble.

\paragraph{Additional baselines.} Two further baselines place the result in
context. Pixel-level majority voting over the five calibrated members
reaches $0.688$ IoU, marginally above both the mean ensemble and the
fitted stack ($0.681$) and still below the best single model.
Zero-shot Segment Anything (SAM\,2.1
Hiera-L \cite{ravi2024sam2}), the foundation model used in our dataset
curation, reaches only $0.410$ IoU in fully automatic mode on the test
images; it becomes competitive only when handed an oracle bounding box
(which lifts it to $0.685$), whereas an oracle centre point reaches only
$0.257$, below its own automatic score; both oracle prompts presuppose the
very localisation the automatic task must solve. A task-specialised model
trained on $199$ images is therefore substantially better than zero-shot
SAM in the automatic setting deployment requires, and remains ahead ($0.707$
against $0.685$) even when SAM is handed the oracle box that no deployed
system could supply.

Table~\ref{tab:meta_compare} compares eight meta-learner families fitted on
the same calibrated out-of-fold cube, holding the cube and a fixed default
operating point constant to isolate the combiner choice. The families
cluster below the best single model ($0.694$): logistic regression leads
($0.681$), the linear SVM and the MLP are close behind ($0.680$ and
$0.678$), and the more flexible tree ensembles overfit the out-of-fold
cube and fall to $0.64$--$0.66$. Because the combiner family is a secondary factor, we
adopt logistic regression for its directly interpretable weights
(Table~\ref{tab:meta_weights}); at the OOF-tuned operating point it
delivers the headline stacked result, which no alternative combiner
improves upon.

\begin{table}[t]
\caption{Meta-learner comparison on the calibrated out-of-fold cube of the
five family champions, at a fixed default operating point to isolate the
combiner choice (per-image test IoU). Every fitted combiner lands below the
best single architecture, and none beats unweighted majority voting, so the
fitting adds nothing here; the flexible tree ensembles overfit the cube. RBF and polynomial SVMs are
fitted on a $15$k-pixel subsample; per-combiner configurations appear in
Appendix~\ref{app:meta}.}
\label{tab:meta_compare}
\centering
\small
\begin{tabular}{@{}lc@{}}
\toprule
Combiner & Test IoU \\
\midrule
Best single architecture & \textbf{0.694} \\
Majority vote (unweighted) & 0.688 \\
Mean ensemble & 0.681 \\
\midrule
Logistic regression (adopted) & 0.681 \\
Linear SVM & 0.680 \\
Multi-layer perceptron & 0.678 \\
XGBoost & 0.659 \\
SVM, RBF kernel & 0.646 \\
Gradient boosting (histogram) & 0.645 \\
Random forest & 0.643 \\
SVM, polynomial kernel & 0.640 \\
\bottomrule
\end{tabular}
\end{table}

The cross-validation design itself has only a secondary effect.
Table~\ref{tab:cv_variants} varies it: doubling the fold count to $K{=}10$
trains each fold model on $90\%$ of the data, and a \emph{nested}
cross-validated stack fits the meta-learner on fold-averaged predictions
that match the test-time regime (Section~\ref{subsec:stacking}). Across all
three designs the calibrated stack fails to reach that design's best
single model (nested values at the default operating point, as the nested
procedure tunes no threshold). The nested design narrows the deficit
furthest, to $0.002$, the expected direction since its meta-learner alone
is fitted on the fold-averaged predictions it meets at test time, yet it
still does not cross the line. The ranking mismatch is therefore real but
not the lever that separates a stack from the best single model
(Figure~\ref{fig:cv_ladder}).

\begin{table}[t]
\caption{Effect of the cross-validation design on the
five-family-champion ensemble (per-image test IoU at the OOF-tuned
operating point). Under every design the calibrated stack sits below
that design's own best single model; raising the fold count lifts the
stack and the best single together rather than closing the gap, and the
nested design narrows the deficit to $0.002$ without eliminating it.}
\label{tab:cv_variants}
\centering
\small
\setlength{\tabcolsep}{4pt}
\begin{tabular}{@{}lccc@{}}
\toprule
CV design & Best single & Mean ens. & Stacked \\
\midrule
$5$-fold & 0.694 & 0.681 & 0.681 \\
$10$-fold & \textbf{0.711} & \textbf{0.687} & \textbf{0.696} \\
Nested $5{\times}5$ & 0.694 & 0.681 & 0.691 \\
\bottomrule
\end{tabular}
\end{table}

\subsection{Hard-negative ablation}
\label{subsec:res_hardneg}

Table~\ref{tab:hardneg_ablation} reports the effect of the synthetic
hard-negative phase. Both rows share an identical principal training
run (Updated SandBoilNet on the real training set plus augmented
synthetic data under the leak-free filter); the second additionally
trains on the confusion-weighted synthetic hard-negative pool of
Section~\ref{subsec:hard_negatives}. This ablation is a single
held-out-split comparison rather than the five-fold protocol, so its
absolute baseline ($0.703$) is not directly comparable to the fold-averaged
figures of Table~\ref{tab:stacking_results}; the controlled quantity is the
within-table difference between the two rows, both run under the identical
split.

\begin{table}[t]
\caption{Hard-negative ablation on the principal run (Updated SandBoilNet on real
$+$ augmented synthetic data). The second row additionally trains on the
confusion-weighted synthetic hard-negative pool of
Section~\ref{subsec:hard_negatives}. The intended effect was a lower
false-positive rate; instead the FP rate rose ($0.018\to0.020$) and IoU
fell ($-0.010$), so the negative phase is excluded from the final model.}
\label{tab:hardneg_ablation}
\centering
\footnotesize
\setlength{\tabcolsep}{3pt}
\begin{tabular}{@{}lccccc@{}}
\toprule
Negative phase & IoU & Dice & FP & Recall & Bal.\,Acc. \\
\midrule
None (real $+$ synth)            & \textbf{0.703} & \textbf{0.826} & \textbf{0.018} & 0.808 & \textbf{0.895} \\
$+$ synthetic, conf-weighted & 0.693 & 0.819 & 0.020 & \textbf{0.809} & 0.894 \\
\bottomrule
\end{tabular}
\end{table}

Contrary to the intent of hard-negative mining, the synthetic
negatives made the model fire \emph{more} rather than less: both
recall and the false-positive rate increased, and
intersection-over-union dropped by $0.023$. The confusion weighting of
\eqref{eq:confusion_score} concentrates training on the negatives the
baseline is most fooled by, but on this small dataset the all-zero-mask
negatives appear to shift the decision threshold toward
over-segmentation rather than suppress confusable structures. We
therefore exclude the negative phase from the final model and report it
as a negative result: useful guidance for future work on synthetic
negatives in this domain.

\subsection{Boundary tolerance and bootstrap uncertainty}
\label{subsec:res_boundary}

Two diagnostics put the champion's headline IoU in context. First, the
fuzzy boil rim makes a strict per-pixel IoU pessimistic: relaxing the
boundary match to a $t{=}2$\,px tolerance band raises the micro-averaged
IoU from $0.671$ at $t{=}0$ to $0.689$, an increase of $0.019$ that is
attributable entirely to sub-annotation-noise rim jitter rather than to
genuine region recovery. The gap is modest, which is itself
informative: the residual error is not dominated by a one- or two-pixel
boundary offset but by whole regions that are missed or hallucinated, so
a tolerance band does not paper over the failure cases.

Both diagnostics are ultimately bounded by the annotations themselves.
Where a boil ends is a judgement call: the disturbed sand grades into a
damp halo with no crisp edge, so some masks enclose that halo and
over-annotate the boil, while others trace only the clearly disturbed
core and under-annotate it. IoU penalises both directions equally, so
this bidirectional label noise puts a ceiling on the attainable score
that no model can cross, and it charges a prediction that is arguably
correct with an error that belongs to the label. The effect falls on
every method compared here alike, so the matched single-split deltas and
fold-level variances we rely on are unaffected; it is the absolute IoU
values, ours and the published baselines' alike, that should be read
with this in mind.

Second, the held-out set is small enough that the point estimate carries
real sampling uncertainty. A non-parametric bootstrap over the held-out
images places a $95\%$ confidence interval of $[0.637, 0.745]$ around the
$0.694$ best single model. The width of this interval, roughly $\pm 0.05$
IoU, is the quantitative form of the test-set-size caveat we raise
throughout: it comfortably brackets the stacked ensemble ($0.681$) and
every stacking variant tested, which is why we report the stack's
shortfall against the best single model as consistent in sign but within
sampling noise; repeating the entire five-fold protocol under three
different fold-partition seeds gives a gap of $-0.007 \pm 0.005$,
negative in every repetition. We therefore lean on \emph{matched}
single-split deltas (Section~\ref{subsec:res_ablation}) and fold-level
variance rather than cross-method point comparisons when differences are
this small; the paired Wilcoxon signed-rank test of
Section~\ref{subsec:res_stacking} ($p=0.45$, median per-image difference
$-0.002$) makes the same point for the stacked-versus-single comparison
directly.

\subsection{Per-architecture meta-learner weight analysis}
\label{subsec:res_weights}

A useful by-product of the linear meta-learner is that its five
weights admit a direct interpretation as the relative
contribution of each backbone to the final stacked output.
Table~\ref{tab:meta_weights} reports those weights alongside the
per-architecture calibration temperatures.

\begin{table}[t]
\caption{Logistic regression meta-learner weights and per-architecture
calibration temperatures for the final ensemble. Weights are normalised so
that their absolute values sum to $1$. A temperature $T_a > 1$ indicates
that the architecture was overconfident and was softened; $T_a < 1$
indicates that it was underconfident and was sharpened.}
\label{tab:meta_weights}
\centering
\footnotesize
\setlength{\tabcolsep}{4pt}
\begin{tabular}{@{}lcc@{}}
\toprule
Base learner & $w_a$ & $T_a$ \\
\midrule
SandBoilNet (ConvNeXt-S$+$scSE) & 0.201 & 1.10 \\
Transformer (SegFormer / MiT-B2) & 0.225 & 1.50 \\
Pyramid (FPN / ConvNeXt-S) & \textbf{0.273} & 1.70 \\
Nested CNN (U-Net++ / EffNet-B4) & 0.152 & 1.20 \\
Atrous (DeepLabV3+ / EffNet-B4) & 0.149 & 1.10 \\
\bottomrule
\end{tabular}
\end{table}

\begin{figure}[t]
\centering
\includegraphics[width=\columnwidth]{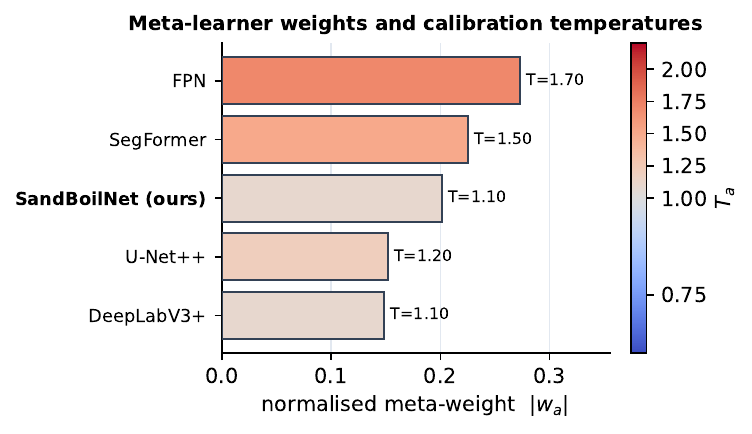}
\caption{Meta-learner weights coloured by calibration temperature: each
backbone's stacked contribution and how much recalibration it required.
Bold labels mark the two SandBoilNet-family backbones we contribute
(SandBoilNet and its Texture-SE variant). The strongest fold-averaged
backbone, SandBoilNet, receives only the fourth-largest weight: the
visible signature of the out-of-fold ranking mismatch analysed in
Section~\ref{subsec:disc_stacking}.}
\label{fig:weights_temp}
\end{figure}

\begin{figure}[t]
\centering
\includegraphics[width=0.92\columnwidth]{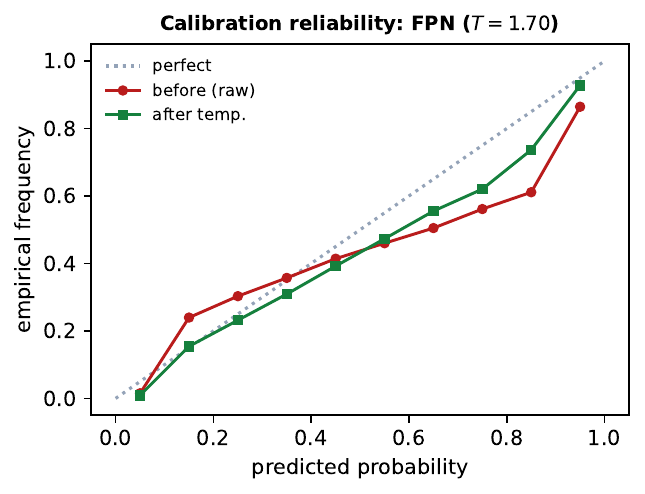}
\caption{Calibration reliability of the most over-confident architecture
before versus after per-architecture temperature scaling: the raw curve
bows below the diagonal (over-confident), and the scaled curve sits close
to it, so the calibrated probabilities are directly combinable. This is
the step that keeps the meta-learner weights of
Table~\ref{tab:meta_weights} proportionate to accuracy rather than to
confidence.}
\label{fig:calibration}
\end{figure}

The pairing of $w_a$ and $T_a$ is informative
(Figure~\ref{fig:weights_temp} plots the two together, and
Figure~\ref{fig:calibration} shows the
reliability curves before and after temperature scaling). A high $|w_a|$
paired with a $T_a$ close to $1$ indicates an architecture whose
probabilities were already well-calibrated and whose stacked contribution
reflects raw accuracy; a high $|w_a|$ paired with a $T_a$ far from $1$
indicates an architecture that required substantial recalibration before
its contribution became proportionate to its accuracy. Refitting the same meta-learner on the \emph{uncalibrated} cube leaves the
stacked test IoU within the bootstrap interval of
Section~\ref{subsec:res_boundary} but redistributes the weights: the
over-confident pyramid and transformer backbones, whose temperatures are
farthest from one, are down-weighted once their probabilities are softened
onto the shared mid-range the meta-learner reads. Calibration is therefore
not an accuracy lever here, and we do not claim it as one; what it buys is
that the weights of Table~\ref{tab:meta_weights} are comparable across
backbones and can be read as relative contributions at all. Notably, the
strongest \emph{single} backbone, SandBoilNet, draws only a middling
weight ($0.201$): fitted on single-fold out-of-fold predictions,
the meta-learner cannot see that this \emph{highest-variance} backbone
(fold-to-fold sd $0.026$ versus $\le 0.021$ for the others;
Table~\ref{tab:fold_variance}) becomes the strongest model once its five
folds are averaged, so it leans on the more stable members. That reweighting is not enough to lift the stacked output
past the best single model, however: the weights redistribute influence
among members whose errors largely coincide, which changes the blend
without adding independent evidence.

\subsection{Qualitative segmentation comparison}
\label{subsec:res_qualitative}

\begin{figure*}[tp]
\centering
\includegraphics[width=0.64\textwidth]{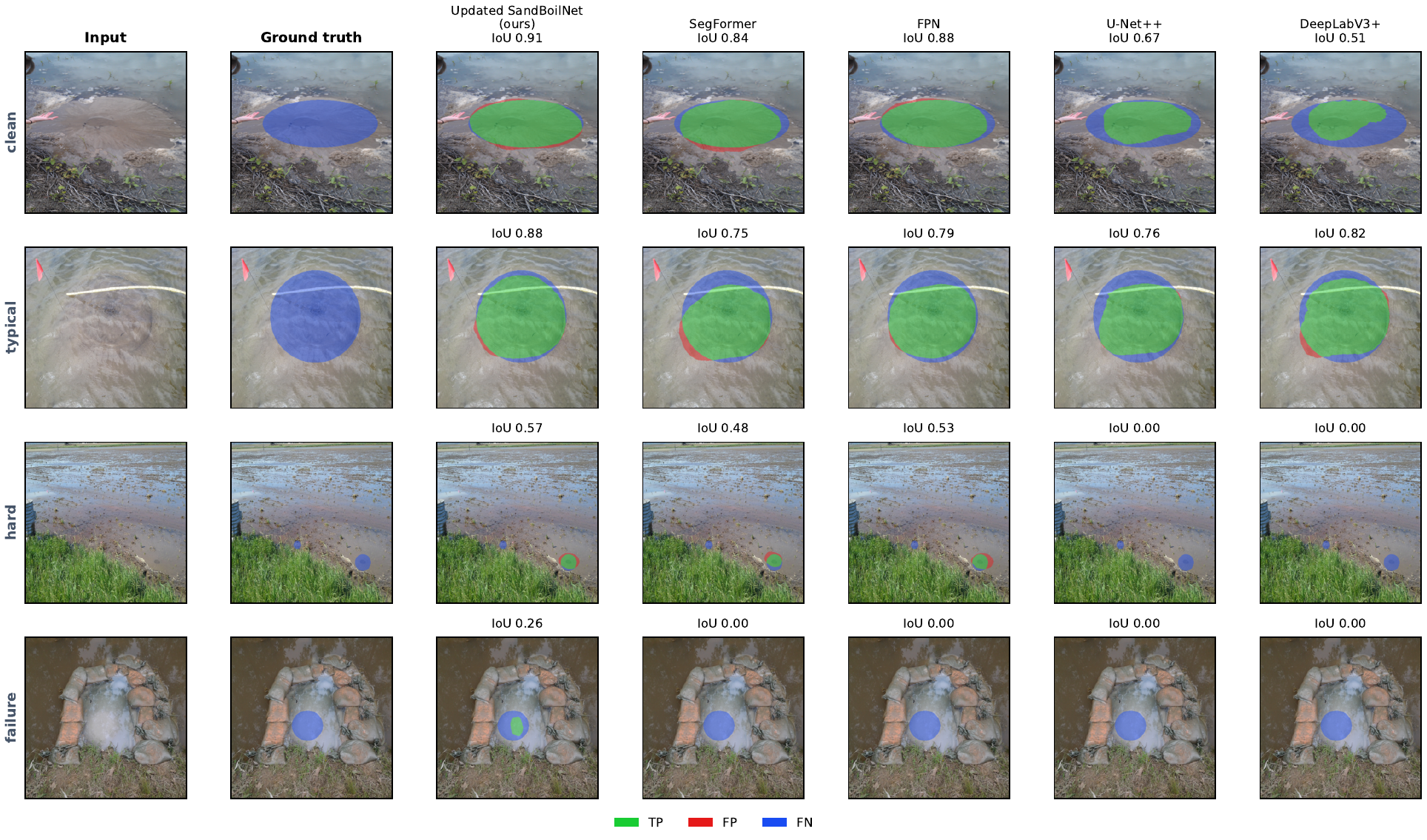}
\caption{Qualitative comparison on held-out real test images, drawn from
those the proposed model leads and spanning its IoU range from a clean
detection (top) to a near-total failure (bottom); full-test-set means for
every architecture are in Table~\ref{tab:stacking_results}. Each model
cell overlays true positives (green), false positives (red), and false
negatives (blue), titled with that image's IoU; the Input and
Ground-truth columns carry no prediction overlay, so blue there is the
annotated mask itself. In the bottom row every model segments a boil-like
wet patch while missing the true boil, the puddle-type confusable
targeted by the hard-negative phase
(Section~\ref{subsec:hard_negatives}).}
\label{fig:segmentation_qualitative}
\end{figure*}

Figure~\ref{fig:segmentation_qualitative} compares the five base
learners' predictions on a fixed set of test images spanning the IoU
range from a clean detection to a failure case. The per-architecture
differences the stacked meta-learner draws on are visible: individual
backbones fragment the sand-boil region or miss its outer rim in
low-contrast or partially occluded scenes, and they disagree most on the
harder rows, which is the error independence a calibrated combiner
exploits. The bottom row is the instructive limit: every backbone fires on
a boil-like wet patch (red) while missing the true boil (blue), a
correlated zero-overlap error that no combiner can repair, and exactly the
puddle-type confusable the hard-negative phase of
Section~\ref{subsec:hard_negatives} targets.

\subsection{Ablations: skip attention, data, curation, resolution}
\label{subsec:res_ablation}

Table~\ref{tab:champion_ablation} isolates the design levers on the
champion backbone (SandBoilNet, ConvNeXt-S$+$scSE) under a single 85/15
split, each averaged over three seeds. None of the alternatives improves
on the plain scSE gate. The texture-sensitive variant (Texture-SE, a
second-order standard-deviation channel gate) is in fact the weakest at
$0.690 \pm 0.012$, $0.016$ below the control; the contrast-gated TC-SE
($-0.001$) and the multi-scale spatial Atrous-SE ($-0.005$) are
indistinguishable from it. A four-fold
increase in input resolution to $1024$\,px also hurts. Attention machinery
beyond the standard scSE gate therefore buys nothing reliable here.

Synthetic training data affects the champion unevenly across
generators. Evaluating each pool separately
(Table~\ref{tab:per_generator}, matched single-split control $0.678$)
spans a $0.046$ swing: the copy-paste compositing baseline ($+0.040$),
the mask-conditioned MaskCN pool ($+0.038$), the merged pool ($+0.028$),
the V2 re-render ($+0.020$) and the V4 inpainting variant ($+0.015$) all
improve the champion, while V1 is neutral ($+0.002$) and V3 slightly
negative ($-0.006$). V3 and V4 share a conditioning stack and differ
only in V4's soft-mask inpainting, which preserves the real boil pixels. Because
identically configured controls vary by up to $0.026$ across runs on
this test set, we read the table as indicative only, and replicated the
two conditions that matter most. Under three-seed replication the
copy-paste pool is neutral ($-0.000$), while the quality-filtered $V_2$
pool (only pairs on which all five folds of a held-out segmenter reach
$0.70$ IoU against the pair's own mask) reaches $0.718 \pm 0.012$ against
the $0.707$ control, a mean gain of $0.012$. That gain is the largest we
observe from any intervention, but its spread is as wide as its mean and
one of the three runs falls below the control, so we report it as
suggestive rather than established. A wider sweep isolating the
generator's guidance scale, over four values with five seeds each on
\emph{unfiltered} pools of the same size, separates the two factors:
guidance scale is immaterial (all settings within $0.002$ of one another),
and without the label-fidelity filter the mean gain over nineteen runs
falls to $0.001$. What little synthetic augmentation buys the strongest
model therefore comes from filtering the pool for label fidelity, not
from the sampling configuration that produced it. Whether or not a given
pool helps, the leak-free filter remains a precondition for reporting
the result at all.

\begin{table}[t]
\caption{Skip-attention design-lever ablation on the champion backbone
(SandBoilNet, ConvNeXt-S$+$scSE), single 85/15 split, mean $\pm$ standard
deviation over three seeds ($\{1,2,42\}$); $\Delta$ is relative to the
scSE control. No variant improves on the plain scSE gate, and the spread
between best and worst is $0.017$, comparable to the seed noise of the
weakest entry. On an earlier, harder split the texture-sensitive gate led
by $0.024$, an advantage that does not survive here.}
\label{tab:champion_ablation}
\centering
\footnotesize
\begin{tabular}{@{}lcc@{}}
\toprule
Skip-attention variant & Test IoU (mean $\pm$ sd) & $\Delta$ \\
\midrule
scSE control & $\mathbf{0.707 \pm 0.003}$ & -- \\
TC-SE (texture $+$ contrast) & $0.705 \pm 0.007$ & $-0.001$ \\
Atrous-SE (multi-scale spatial) & $0.701 \pm 0.007$ & $-0.005$ \\
Texture-SE (std channel gate) & $0.690 \pm 0.012$ & $-0.016$ \\
\bottomrule
\end{tabular}
\end{table}

\begin{table}[t]
\caption{Per-generator synthetic-pool ablation on the champion, single held-out
split under the leak-free per-fold filter; $\Delta$ is relative to the
matched real-only control of the same run ($0.678$). The $V$ presets
differ in their multi-ControlNet conditioning; MaskCN is the
mask-conditioned generator of Section~\ref{subsec:maskcn}, evaluated on a
leak-safe pool (Appendix~\ref{app:maskcn}); copy-paste is the
compositing baseline of Appendix~\ref{app:copypaste}. Identically
configured controls vary by up to $0.026$ on this test set, so individual
deltas are indicative rather than exact.}
\label{tab:per_generator}
\centering
\footnotesize
\setlength{\tabcolsep}{3pt}
\begin{tabular}{@{}llcc@{}}
\toprule
Generator & ControlNet conditioning & IoU & $\Delta$ \\
\midrule
Real only (control) & -- & 0.678 & -- \\
Copy-paste & compositing baseline & \textbf{0.718} & $\mathbf{+0.040}$ \\
MaskCN & mask-conditioned ControlNet & 0.716 & $+0.038$ \\
Merged pool & all presets combined & 0.706 & $+0.028$ \\
V2 & Canny\,$+$\,Normal\,$+$\,HED & 0.698 & $+0.020$ \\
V4 & as $V_3$, $+$\,soft-mask inpaint & 0.693 & $+0.015$ \\
V1 & Canny\,$+$\,Depth\,$+$\,HED & 0.680 & $+0.002$ \\
V3 & HED\,$+$\,Normal\,$+$\,IPA & 0.672 & $-0.006$ \\
\bottomrule
\end{tabular}
\end{table}


\subsection{Cross-dataset check on DeepCrack}
\label{subsec:res_deepcrack}

To verify that the architecture ranking is not an artefact of one
dataset, we retrain all five base learners from scratch on DeepCrack
\cite{liu2019deepcrack} ($300$ train / $237$ test crack images), using
the identical pipeline, augmentation and per-architecture schedules,
with no synthetic data and no stacking. Over the same three seeds the
Updated SandBoilNet again leads
(Table~\ref{tab:deepcrack}), indicating that the modernised
ConvNeXt-S$+$scSE configuration transfers beyond sand boils to a second
scarce-data infrastructure-defect task.

\begin{table}[t]
\caption{Cross-dataset check: the five base learners retrained on
DeepCrack ($300$/$237$), mean $\pm$ sd of test IoU over seeds
$\{1,2,42\}$; same pipeline as the sand-boil experiments, real data
only.}
\label{tab:deepcrack}
\centering
\footnotesize
\setlength{\tabcolsep}{4pt}
\begin{tabular}{@{}lc@{}}
\toprule
Base learner & Test IoU (mean $\pm$ sd) \\
\midrule
Updated SandBoilNet (ConvNeXt-S$+$scSE) & $\mathbf{0.737 \pm 0.006}$ \\
U-Net++ / EffNet-B4 & $0.733 \pm 0.014$ \\
FPN / ConvNeXt-S & $0.726 \pm 0.003$ \\
DeepLabV3+ / EffNet-B4 & $0.711 \pm 0.012$ \\
SegFormer / MiT-B2 & $0.707 \pm 0.009$ \\
\bottomrule
\end{tabular}
\end{table}

\subsection{Computational cost}
\label{subsec:res_cost}

The full five-fold, five-architecture training completes in
$\approx$1.7\,h on a single NVIDIA H100 NVL, and stacked inference is
sub-second per image; a per-phase breakdown is given in
Appendix~\ref{app:cost} (Table~\ref{tab:cost}). The upstream generation
costs are reported in the companion paper \cite{thapa2026diffusion}.

\section{Discussion}
\label{sec:discussion}

Section~\ref{sec:results} supports a compact reading. The strongest
single backbone reaches $0.707$ IoU with an scSE skip-attention gate, and
the texture-sensitive variant we expected to lead trails it by $0.016$. A
calibrated stack over five architectures reaches $0.681$ without
improving on the best single model, and the synthetic hard-negative phase
costs a further $0.010$ IoU. We examine each of these below, since a
protocol built so that such measurements can be believed is the central
contribution.

\subsection{Cross-validated stacking: a trustworthy protocol without an
ensemble gain}
\label{subsec:disc_stacking}

The weighted-average ensemble of our previous work \cite{thapa2025thesis}
fits its weights on the same split later used to report accuracy, a mild
but real leakage. Cross-validated stacking removes it, because every
out-of-fold prediction comes from a model that did not observe the
example and the meta-learner is fitted on those predictions alone,
leaving both the estimate and the weights unbiased. That guarantee holds
independently of the size of any gain.

Under this measurement the ensemble does not win. The calibrated stack
reaches $0.681$ against $0.694$ for the best single network, and a plain
mean ensemble matches it. The outcome is not an artefact of the combiner,
since eight meta-learners spanning linear, kernel, neural, and tree
families (Table~\ref{tab:meta_compare}) land at or below the best single
model. Nor is it an artefact of the protocol, since the leakage audit
confirms that the meta-learner observes out-of-fold predictions
exclusively.

Member correlation explains the result. The members' per-pixel error maps
correlate at $0.894$ averaged over all pairs, and the tightest pairs,
U-Net with FPN ($0.931$) and U-Net++ with DeepLabV3+ ($0.929$), are those
that share an encoder. Correlation therefore tracks the encoder rather
than the decoder: five decoders rest on only three feature extractors, so
the ensemble carries less independence than its member count suggests,
and five strong backbones trained on the same $199$ images from a common
initialisation fail on the same boundaries.

The information is nonetheless present, since a per-pixel oracle
selecting the best member at each location reaches $0.837$ IoU, so the
ceiling is set by what the combiner can reach rather than by what the
members encode. Replacing the FPN and DeepLabV3+ encoders with ResNet50
and ResNet101, giving five distinct feature extractors, narrows the
shortfall from $-0.013$ to $-0.008$ without reversing it ($0.684$ against
$0.693$). Decorrelation thus recovers about a third of the deficit, the
remainder being imposed by the shared training images rather than by the
choice of backbones. Once one architecture is clearly strongest, a
combiner obliged to weight all five is handicapped against selecting it.
Two secondary effects deserve record. The out-of-fold ranking mismatch,
whereby the meta-learner observes single-fold predictions but is applied
to fold averages at test time, accounts for most of the measured
shortfall: nested cross-validation lifts the stacked score from $0.681$ to
$0.691$, narrowing the deficit to $0.002$. That shift is smaller than the
$0.026$ run-to-run spread of this test set, so we do not present it as
established, and the nested stack still does not exceed its best member.
Calibration and the operating point likewise matter more than the combiner
family.

\subsection{Leak-free provenance and calibration}
\label{subsec:disc_source_tagging}

The filter of Section~\ref{subsec:leak_free} reduces to one set operation
per fold, yet its consequences for validity are out of proportion to its
size. Without it a synthetic image generated from a real parent can train
a model whose validation fold contains that parent, and even a visually
distinct render carries the parent's geometry, lighting, and texture. The
\texttt{source\_image\_id} field written by the upstream pipeline
\cite{thapa2026diffusion} closes this path, and we recommend it for any
synthetic pipeline evaluated under cross-validation. MaskCN
(Section~\ref{subsec:maskcn}) offers the complementary escape, since a
record synthesised into a target mask has no parent to leak and renders
Eq.~\eqref{eq:leakfree} vacuous. Per-architecture temperature calibration
contributes one scalar per backbone but is not marginal: without it the
most overconfident architecture dominates the meta-learner regardless of
its accuracy, and with it the fitted weights become interpretable as a
record of each architecture's contribution.

\subsection{Confusion-score mining and why it did not help here}
\label{subsec:disc_confusion}

The classical recipe \cite{shrivastava2016hardnegative} mines high-loss
examples from an unlabelled real pool, whereas we mine high-confusion
examples from a controllable synthetic one, weighting each by
Eq.~\eqref{eq:confusion_score} so that the negative-phase gradient
concentrates on the examples that most mislead the baseline. That
mechanism did not help on this dataset. The negative phase lowered test
IoU from $0.703$ to $0.693$ and raised the false-positive rate from
$0.018$ to $0.020$ while leaving recall essentially unchanged at $0.808$
against $0.809$ (Table~\ref{tab:hardneg_ablation}), so it bought no
recall and cost both precision and accuracy. The most plausible cause is
open-loop scoring, since the negatives are scored once by the public
baseline and never re-scored against the model being trained, leaving the
pool unadapted to the residual errors of the modernised backbone. A
closed-loop variant is the natural next step, but on the present evidence
we do not recommend the open-loop phase as a default.

\subsection{Limitations, threats to validity, and ethical considerations}
\label{subsec:disc_failures}

Several limitations qualify these conclusions. The held-out test set is
small in absolute terms, so single-digit differences are not always
distinguishable from variation across splits and a paired comparison is
unlikely to resolve improvements smaller than a few IoU points; we
mitigate this by averaging fold predictions and fixing the partition and
seed cascade. Predictions still degrade where a boil is obscured by
reflective standing water, which breaks the texture cues the network
relies on, and a two-pass approach that first predicts reflection regions
is left to future work. Sand boils also co-occur with seepage or cracks
within one frame, which the per-class models used here do not capture, so
a shared multi-class head is a natural extension. The confusion score
depends on the publicly released baseline, so a category on which that
checkpoint silently fails is scored as easy when it is in fact hard.
Finally, the synthetic data derives from a single instance of the
upstream pipeline \cite{thapa2026diffusion}, so gains observed under
combined real and synthetic training may partly reflect its stylistic
biases; an independent generator on a disjoint reference set would
constitute a stronger external control.

The pipeline is intended to support trained inspectors rather than
replace them. Synthetic data introduces a risk of overconfidence, since
high in-distribution accuracy need not transfer to field conditions
absent from the reference set, and the protocol reports fold-level
variance so that deployment can expose that uncertainty rather than
collapse it to a binary decision. Because a false negative during a
high-water event can delay intervention long enough to permit a breach,
deployment should err toward false positives verified by a human
inspector.

\section{Conclusion and Future Work}
\label{sec:conclusion}

We have presented a segmentation methodology for pixel-level sand boil
detection that combines five encoder--decoder backbones through a
calibrated cross-validated stacking ensemble and strengthens its
rejection of visually confusable non-defects through a
confusion-score-mined synthetic hard-negative phase. The framework closes
two leakage paths the defect-segmentation literature has not consistently
addressed: the meta-learner is fitted on out-of-fold predictions alone,
so its weights are unbiased, and every synthetic record carries a
\texttt{source\_image\_id} pointer to its real parent, so a per-fold
filter keeps synthetic descendants of held-out images out of the
corresponding training folds. Per-architecture temperature calibration
makes the backbones comparable before stacking and leaves the linear
meta-learner's weights readable as each backbone's relative contribution.

On a held-out real sand-boil test set the Updated SandBoilNet reaches
$0.707$ intersection-over-union with an scSE skip-attention gate. A
calibrated stack over five architectures reaches $0.681$ and does not
exceed the best single model, an outcome eight meta-learner families
reproduce, so the shortfall is a property of the members rather than of
the combiner. A controlled ablation clarifies what does and does not
drive accuracy here: five strong backbones trained on the same scarce
data are too correlated for a combiner to exploit; the texture-sensitive
gate that led on our earlier split is now the weakest of four variants;
synthetic augmentation repays the strongest model when the pool is
filtered for label fidelity, though not through the sampling
configuration that generated it; and neither the hard-negative phase nor
a four-fold resolution increase improves on it. Establishing which
additions transfer to a scarce-data, fuzzy-boundary defect and which do
not is a deliberate complement to the positive gains. The synthetic
data is produced by the pipeline of our companion paper
\cite{thapa2026diffusion} and consumed here as an upstream module.

Several extensions remain open, each developed in
Section~\ref{sec:discussion}. The hard-negative loop is currently
one-directional, and a closed-loop variant would re-score with the
current stacked ensemble after each round, while the symmetric version on
the positive pool would steer synthesis toward the conditions the model
still fails on, making the generator an active learner rather than a
fixed data source. MaskCN (Section~\ref{subsec:maskcn}) is the most
immediate lever, supplying registered labels at zero annotation cost with
no parent-image filter required, though folding curated MaskCN singles
into the augmentation mixture at a fidelity that helps the champion
remains open. Since our comparison isolates the encoder as the dominant
factor, transformer-based dense-prediction decoders and depth-aware
architectures are a natural next step for a boundary this diffuse.
Beyond that, a shared multi-class head would capture sand boils
co-occurring with seepage or cracks, and a public levee-defect benchmark
with paired real and synthetic data, source tags, and a held-out split
would let the community compare methods directly; that dataset and the
deployed inspection workflow are the subject of the third paper in this
series.

\section*{Code and Data Availability}
\label{sec:availability}

The complete segmentation and stacking framework is released alongside
this paper, comprising the per-architecture training driver, the
cross-validated stacking driver, and the ablation, baseline, and
hard-negative scripts that produce every table in
Section~\ref{sec:results}. The artefact also includes the twenty-five
fold checkpoints (five architectures $\times$ five folds), the
out-of-fold probability cubes, the fitted temperature scalars and
meta-learner weights of Table~\ref{tab:meta_weights}, and three manifests
that make the evaluation auditable rather than merely repeatable: the
per-fold \texttt{source\_image\_id} admission list, which records exactly
which synthetic records entered each fold's training pool and so allows
the leak-free filter to be verified independently; the confusion-scored
hard-negative pool manifest; and the mask-source manifest of the
leak-safe MaskCN evaluation pool (Appendix~\ref{app:maskcn}). With the
seeds recorded in Section~\ref{subsec:reproducibility_exp}, every
numerical entry in Section~\ref{sec:results} re-derives from a single
command. On acceptance the artefact will be published at
\url{https://github.com/padam56/sandboil-leakfree-stacking} and archived
under a versioned Zenodo DOI.

The upstream generative components are released with the companion
generation paper \cite{thapa2026diffusion}, and the \textsc{Convex Hull
Annotator} used in curation is distributed as a standalone,
class-agnostic library \cite{thapa_chull_annotator}. The real imagery
derives from the U.S.\ Army Corps of Engineers levee-inspection archive
\cite{usace_levee_manual}, whose terms of use preclude redistribution of
the underlying photographs; the pixel-level binary masks produced by the
authors will be released under CC~BY~4.0 alongside the code, on the same
terms as the group's earlier public levee-defect releases
\cite{thapa2024dataport, thapa2025dataport2}.

\section*{Conflicts of Interest}

The authors declare no competing financial or non-financial interests.
This work was carried out in an academic research setting and is derived
from the corresponding author's Master's thesis \cite{thapa2025thesis}.
All authors have read and approved the final manuscript and agree to its
submission.

\section*{Acknowledgments}
\label{sec:acks}

This research was supported by the U.S.\ Army Corps of Engineers
(USACE), under Contract No.~W912HZ-23-2-0004. The authors thank the
USACE for making the levee-inspection imagery archive
\cite{usace_levee_manual} available for academic research and
acknowledge Stability AI, the Diffusers library, the ControlNet
authors, and the IP-Adapter authors for their open-source models and
software. This work was conducted on the institutional GPU cluster at
LSU New Orleans. The views and conclusions expressed in this paper are
those of the authors and do not necessarily reflect the official
policies or positions of the USACE.

\FloatBarrier                      
\appendices

\section{Meta-learner configurations and the combiner-by-design grid}
\label{app:meta}

Table~\ref{tab:meta_compare} in the main text reports the eight
per-pixel meta-learners under the primary five-fold design. This
appendix records the exact configurations and the full
combiner-by-design grid, so that every number is reproducible from the
released sweep script with fixed seeds.

All combiners consume the same input: the per-pixel five-dimensional
vector of temperature-calibrated architecture probabilities
(Section~\ref{subsec:stacking}). Each is fitted on a fixed-seed random
pixel subsample of the out-of-fold cube: $2$M pixels for logistic
regression (\texttt{max\_iter}$=500$, $C{=}1$), the linear SVM
($C{=}1$), histogram gradient boosting (library defaults), and XGBoost
($300$ trees, histogram method); $1$M pixels for the MLP (two hidden
layers, $32$ and $16$ units, early stopping) and the random forest
($300$ trees); and $15$k pixels for the RBF and polynomial
(degree-$3$) SVMs, whose kernel fits do not scale further. Margin
classifiers are thresholded at a decision score of $0$, probabilistic
ones at $0.5$. Test-time inputs are the calibrated fold-averaged test
probabilities of the corresponding design.

\begin{table}[!ht]
\caption{Combiner-by-design grid: per-image test IoU of the eight meta-learners
under the $5$-fold and $10$-fold designs on the five-family-champion cube
at the default operating point, with each design's best-single and
mean-ensemble anchors. The $5$-fold column reproduces
Table~\ref{tab:meta_compare} for comparison with $K{=}10$. No combiner
under either design exceeds the best single model on this base cube.}
\label{tab:meta_grid}
\centering
\small
\setlength{\tabcolsep}{5pt}
\begin{tabular}{@{}lcc@{}}
\toprule
Combiner & $5$-fold & $10$-fold \\
\midrule
Best single architecture & \textbf{0.694} & \textbf{0.711} \\
Mean ensemble & 0.681 & 0.687 \\
\midrule
Logistic regression (adopted) & 0.681 & 0.696 \\
Linear SVM & 0.680 & 0.694 \\
Multi-layer perceptron & 0.678 & 0.681 \\
Gradient boosting (histogram) & 0.645 & 0.687 \\
SVM, RBF kernel & 0.646 & 0.666 \\
XGBoost & 0.659 & 0.657 \\
SVM, polynomial kernel & 0.640 & 0.649 \\
Random forest & 0.643 & 0.633 \\
\bottomrule
\end{tabular}
\end{table}

Two readings of Table~\ref{tab:meta_grid} matter for the main text.
First, the ordering shifts systematically with the design: the combiners
that overfit the single-fold out-of-fold signal at $K{=}5$ (the RBF SVM,
gradient boosting) gain IoU at $K{=}10$, while the linear combiners barely
move, direct evidence for the ranking-mismatch effect of
Section~\ref{subsec:res_stacking}. Second, on this base cube at the default
operating point no combiner under either design reaches the best single
model; the main text's finding (Table~\ref{tab:stacking_results}) is that
no combiner reaches the best single model, and this grid shows that
conclusion is driven neither by the combiner family nor by the
cross-validation design. Adopting a flexible combiner on the strength of
its test score would violate the selection discipline this paper argues
for, which is why we fix logistic regression for its interpretable weights.

\begin{table}[!ht]
\caption{Per-architecture fold-level variance for the five family champions of
Table~\ref{tab:arch_search}: held-out test IoU of each individual fold
model, before within-architecture averaging. SandBoilNet is the
highest-variance backbone, which is why the meta-learner
(Table~\ref{tab:meta_weights}), fitted on single-fold out-of-fold
predictions, gives it a modest weight.}
\label{tab:fold_variance}
\centering
\footnotesize
\setlength{\tabcolsep}{4pt}
\begin{tabular}{@{}lcc@{}}
\toprule
Base learner & Fold IoU (mean $\pm$ sd) & Range \\
\midrule
SandBoilNet & $0.630 \pm 0.026$ & $0.600$--$0.672$ \\
FPN & $0.633 \pm 0.017$ & $0.612$--$0.663$ \\
SegFormer & $0.629 \pm 0.021$ & $0.608$--$0.660$ \\
U-Net++ & $0.634 \pm 0.020$ & $0.610$--$0.670$ \\
DeepLabV3+ & $0.620 \pm 0.015$ & $0.604$--$0.647$ \\
\bottomrule
\end{tabular}
\end{table}

\section{Leak-safe construction of the MaskCN evaluation pool}
\label{app:maskcn}

The MaskCN generator of Section~\ref{subsec:maskcn} renders a fresh
sand boil into a conditioning mask drawn from a bank of $41$ real
annotation masks. Although MaskCN records are parentless in image
space, the conditioning mask itself carries the label geometry of its
source image: a sample conditioned on the mask of a \emph{held-out}
image would place that image's exact ground-truth shape in the
training set. Eight of the $41$ bank masks derive from test-set
images: \mbox{img\,(9)}, (14), (17), (35), (92), (119), (172), and
(176). An unfiltered pool therefore leaks test-label geometry.

The evaluation pool of Table~\ref{tab:per_generator} therefore keeps
only samples conditioned on the $33$ training-set mask sources:
of the $585$ paired candidates, $111$ test-source samples are
excluded and $474$ are retained, each carrying its mask-source
identifier in the pool manifest. Training then follows the identical
per-generator protocol (champion backbone, seed $42$, matched
single-split control). No result from any unfiltered MaskCN pool is
reported anywhere in this paper.

\begin{figure*}[t]
\centering
\includegraphics[width=0.66\textwidth]{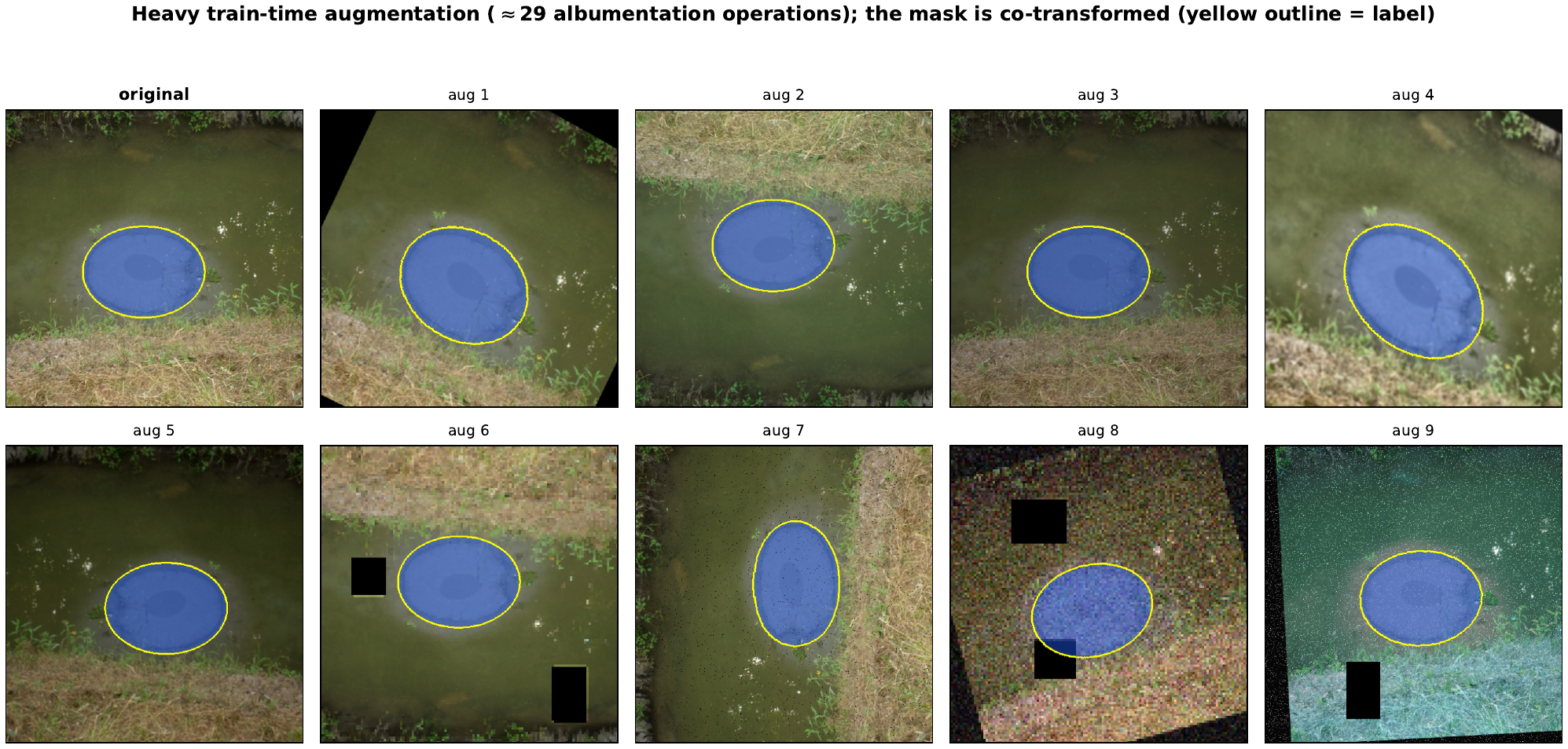}
\caption{The heavy train-time augmentation pipeline (the $\approx$29
Albumentations operations of Table~\ref{tab:aug_ops}) applied to one
training image, with the mask co-transformed (yellow outline marks the
propagated label). Panels \textit{aug 1}--\textit{aug 9} are nine
independent draws, each a random subset of the operations rather than any
single named transform, selected from a fixed seed range so that the nine
between them exercise all twenty-nine operations. Validation and test
images are never augmented.}
\label{fig:aug_showcase}
\end{figure*}

\section{Copy-paste augmentation baseline}
\label{app:copypaste}

As a simple augmentation reference we also trained the champion backbone
with a copy-paste pool \cite{ghiasi2021copypaste}, in which real boil
instances, cut out with their masks, are pasted onto other training images.
Under the same three-seed protocol it reaches $0.706$ IoU, on par with
the real-only champion ($0.707$),
consistent with the known strength of copy-paste for instance
segmentation. We report it for completeness; the paper's focus is
generative augmentation for its controllability and procedural scene
diversity, which copy-paste cannot provide.

\section{Computational cost}
\label{app:cost}

Table~\ref{tab:cost} breaks the wall-clock cost of the segmentation
pipeline down by phase on a single H100 NVL, complementing the summary in
Section~\ref{subsec:res_cost}: the full five-fold, five-architecture
training completes in about $2$\,h and stacked inference is sub-second
per image.

\begin{table}[!ht]
\caption{Wall-clock cost of the segmentation pipeline on a single NVIDIA
H100 NVL GPU. The upstream generation costs (DreamBooth fine-tuning,
augmented dataset synthesis, hard-negative pool generation) are reported
in the companion paper \cite{thapa2026diffusion}.}
\label{tab:cost}
\centering
\footnotesize
\setlength{\tabcolsep}{4pt}
\begin{tabular}{@{}lc@{}}
\toprule
Phase & H100 \\
\midrule
Single-architecture training, principal phase & $\approx$9\,min \\
Five-fold CV, 5 architectures (25 trainings) & $\approx$2\,h \\
Out-of-fold prediction collection & $<$1\,min \\
Temperature calibration (per architecture) & $<$1\,s \\
Meta-learner training (logistic regression) & $<$10\,s \\
Stacked inference, per image & $<$1\,s \\
\bottomrule
\end{tabular}
\end{table}

\section{Per-architecture training configuration}
\label{app:training_config}

Table~\ref{tab:arch_config} records the exact optimiser and schedule
settings behind the architecture search and the cross-validation runs, so
every training reproduces from the released configuration. All five base
learners share the combined BCE-plus-Dice objective, the AdamW optimiser,
a batch size of $8$ at $512\times512$, automatic mixed-precision training,
a five-epoch linear warmup into cosine annealing, and early stopping on
validation IoU (patience $30$, with the best-IoU checkpoint restored for
testing). They differ only in the locked encoder--decoder pairing and in
the architecture-appropriate learning rate and weight decay: the
transformer takes the smallest learning rate and the heaviest weight decay
to stabilise its global attention, while the two ConvNeXt-backed models
sit between the transformer and the EfficientNet CNN baselines.

\begin{table}[!ht]
\caption{Per-architecture training configuration. All models share the
BCE-plus-Dice loss, AdamW, batch size $8$, $512\times512$ input,
mixed-precision, a five-epoch warmup into cosine annealing, and early
stopping (patience $30$); they differ only in the encoder--decoder pairing
and the architecture-appropriate learning rate and weight decay.}
\label{tab:arch_config}
\centering
\footnotesize
\setlength{\tabcolsep}{4pt}
\begin{tabular}{@{}llcc@{}}
\toprule
Base learner (decoder) & Encoder & LR & Wt.\ decay \\
\midrule
SandBoilNet (U-Net$+$scSE) & ConvNeXt-S & $8\mathrm{e}{-5}$ & $1\mathrm{e}{-4}$ \\
SegFormer & MiT-B2 & $6\mathrm{e}{-5}$ & $1\mathrm{e}{-2}$ \\
FPN & ConvNeXt-S & $8\mathrm{e}{-5}$ & $1\mathrm{e}{-4}$ \\
U-Net$++$ & EfficientNet-B4 & $1\mathrm{e}{-4}$ & $1\mathrm{e}{-4}$ \\
DeepLabV3$+$ & EfficientNet-B4 & $1\mathrm{e}{-4}$ & $1\mathrm{e}{-4}$ \\
\bottomrule
\end{tabular}
\end{table}

\section{Augmentation pipeline}
\label{app:aug_pipeline}

The heavy train-time pipeline of Section~\ref{subsec:seg_protocol}
(illustrated in Figure~\ref{fig:aug_showcase}) applies the twenty-nine
Albumentations operations of Table~\ref{tab:aug_ops}. The geometric
transforms are applied jointly to image and mask so the propagated label
stays pixel-aligned through every warp; the photometric, noise, blur, and
codec transforms act on the image alone. Operations that would otherwise
compound destructively are wrapped in mutually exclusive \texttt{OneOf}
blocks, so that, for example, at most one of the four blur operators
fires on any given sample. Validation and test images receive only the
resize-and-normalise tail, never augmentation.

\begin{table}[!ht]
\caption{The twenty-nine train-time Albumentations operations
(Section~\ref{subsec:seg_protocol}), grouped by stage, with the
per-operation or per-block application probability~$p$. A bracketed group
is a \texttt{OneOf} block: $p$ selects the block, then one member fires
uniformly at random.}
\label{tab:aug_ops}
\centering
\footnotesize
\setlength{\tabcolsep}{4pt}
\begin{tabular}{@{}llc@{}}
\toprule
Stage & Operation & $p$ \\
\midrule
Geometric & horizontal flip & $0.50$ \\
          & vertical flip & $0.50$ \\
          & random $90^\circ$ rotate & $0.20$ \\
          & transpose & $0.15$ \\
          & rotate ($\pm35^\circ$) & $0.35$ \\
          & affine (scale/translate/rotate) & $0.30$ \\
          & perspective & $0.15$ \\
          & \{elastic, grid, optical distortion\} & $0.25$ \\
\midrule
Photometric & brightness/contrast & $0.35$ \\
            & \{colour-jitter, HSV, RGB-shift, gamma\} & $0.35$ \\
            & channel shuffle & $0.10$ \\
            & CLAHE & $0.20$ \\
\midrule
Noise / dropout & \{Gaussian, ISO, multiplicative\} & $0.20$ \\
                & pixel dropout & $0.10$ \\
                & coarse dropout & $0.12$ \\
\midrule
Blur / codec & \{blur, Gaussian, median, motion\} & $0.18$ \\
             & sharpen & $0.12$ \\
             & JPEG compression & $0.12$ \\
             & downscale & $0.10$ \\
\bottomrule
\end{tabular}
\end{table}

\section{Applying the protocol: a checklist}
\label{app:checklist}

The protocol of Section~\ref{subsec:leak_free} is a discipline rather than
an architecture, so it transfers to any pipeline that mixes generated and
real images under cross-validation. Three requirements carry it.

\emph{1) Record provenance at generation time.} Every synthetic record
must carry the identifier of its real parent
(\texttt{source\_image\_id}) at the moment it is written to disk.
Parentage cannot be recovered afterwards, because a generator conditioned
on image $x$ can emit a scene resembling image $y$ more closely than $x$.
Records without a parent are safe only when they have none by
construction, as in MaskCN (Section~\ref{subsec:maskcn}).

\emph{2) Filter per fold, never once globally.} The exclusion in
Eq.~\eqref{eq:leakfree} is evaluated against the current fold's held-out
parents and therefore yields a different synthetic subset for each of the
$K$ folds. Filtering once against a fixed validation split and reusing the
survivors reintroduces the leak for $K-1$ folds, and does so silently,
since the run completes and only the reported accuracy is wrong.

\emph{3) Fit every tuned quantity on out-of-fold predictions alone.} The
per-architecture temperature of Eq.~\eqref{eq:tempscale} and the decision
threshold $\tau$ are fitted quantities no less than the meta-learner
weights, and fitting either on the test split restores exactly the bias
the design removes.

Two assertions in the fold-construction driver catch most violations: that
each fold's training pool never intersects the held-out parents or their
descendants, and that the out-of-fold cube holds one prediction per real
image per architecture, each from the fold model that did not train on it.
An off-by-one in the fold index is invisible in aggregate accuracy yet
corrupts every weight fitted downstream. Finally, the protocol makes a
reported gain trustworthy without creating one: here it returned no gain
at all, because the members were too correlated to combine
(Section~\ref{sec:discussion}).

\FloatBarrier                      

\let\origthebibliography\thebibliography
\renewcommand{\thebibliography}[1]{%
  \origthebibliography{#1}%
  \setlength{\itemsep}{-0.7pt}\setlength{\parsep}{0pt}}
\bibliographystyle{IEEEtran}
\bibliography{references}

\end{document}